\documentclass[10pt,twocolumn,letterpaper]{article}

\usepackage{cvpr}
\usepackage{times}
\usepackage{epsfig}
\usepackage{graphicx}
\usepackage{amsmath}
\usepackage{amssymb}
\usepackage{booktabs}
\usepackage{bm}
\usepackage{caption}
\usepackage{subcaption}
\usepackage[breaklinks=true,bookmarks=false]{hyperref}

\cvprfinalcopy

\begin{document}

\title{Deeply learned face representations are sparse, selective, and robust}

\author{Yi Sun$^{1}$~~~~~~~~~~~~~~~~~~~~~Xiaogang Wang$^{2,3}$~~~~~~~~~~~~~~~~~~~~~Xiaoou Tang$^{1,3}$\\
{\small $^1$Department of Information Engineering, The Chinese University of Hong Kong}\\
{\small $^2$Department of Electronic Engineering, The Chinese University of Hong Kong}\\
{\small $^3$Shenzhen Institutes of Advanced Technology, Chinese Academy of Sciences}\\
{\tt\small sy011@ie.cuhk.edu.hk~~~~~xgwang@ee.cuhk.edu.hk~~~~~xtang@ie.cuhk.edu.hk}}

\maketitle

\begin{abstract}
   This paper designs a high-performance deep convolutional network (DeepID2+) for face recognition. It is learned with the identification-verification supervisory signal. By increasing the dimension of hidden representations and adding  supervision to early convolutional layers, DeepID2+ achieves new state-of-the-art on LFW  and YouTube Faces benchmarks.

    Through empirical studies, we have discovered three properties of its deep neural activations critical for the high performance: sparsity, selectiveness and robustness.  (1) It is observed that neural activations are moderately sparse. Moderate sparsity maximizes the discriminative power of the deep net as well as the distance between images. It is surprising that DeepID2+ still can achieve high recognition accuracy even after the neural responses are binarized. (2) Its neurons in higher layers are highly selective to identities and identity-related attributes. We can identify different subsets of neurons which are either constantly excited or inhibited when different identities or attributes are present.  Although  DeepID2+  is not taught to distinguish attributes during training, it has implicitly learned such high-level concepts. (3) It is much more robust to occlusions, although occlusion patterns are not included in the training set.
\end{abstract}

\section{Introduction}

Face recognition  achieved great progress thanks to extensive research effort devoted to this area \cite{wright2009,taigman2009,kumar2009,huang2011,berg2012,chen2012,huang2012,simonyan2013,chen2013,cao2013,sun2013b,taigman2014a,sun2014a,taigman2014b,sun2014b}. While pursuing higher performance is a central topic, understanding the mechanisms behind it is equally important. When deep neural networks begin to approach  human on challenging face benchmarks \cite{taigman2014a,sun2014a,taigman2014b,sun2014b} such as LFW \cite{huang2007a}, people are eager to know what has been learned by these neurons and how such high performance is achieved. In cognitive science, there are a lot of studies  \cite{tsao2008} on analyzing the mechanisms of face processing of neurons in  visual cortex. Inspired by those works,  we analyze the behaviours of neurons in artificial neural networks in a attempt to explain  face recognition process in deep nets, what information is encoded in  neurons, and how robust they are to  corruptions.

\begin{figure}[t]
\begin{center}
\includegraphics[width=1.0\linewidth]{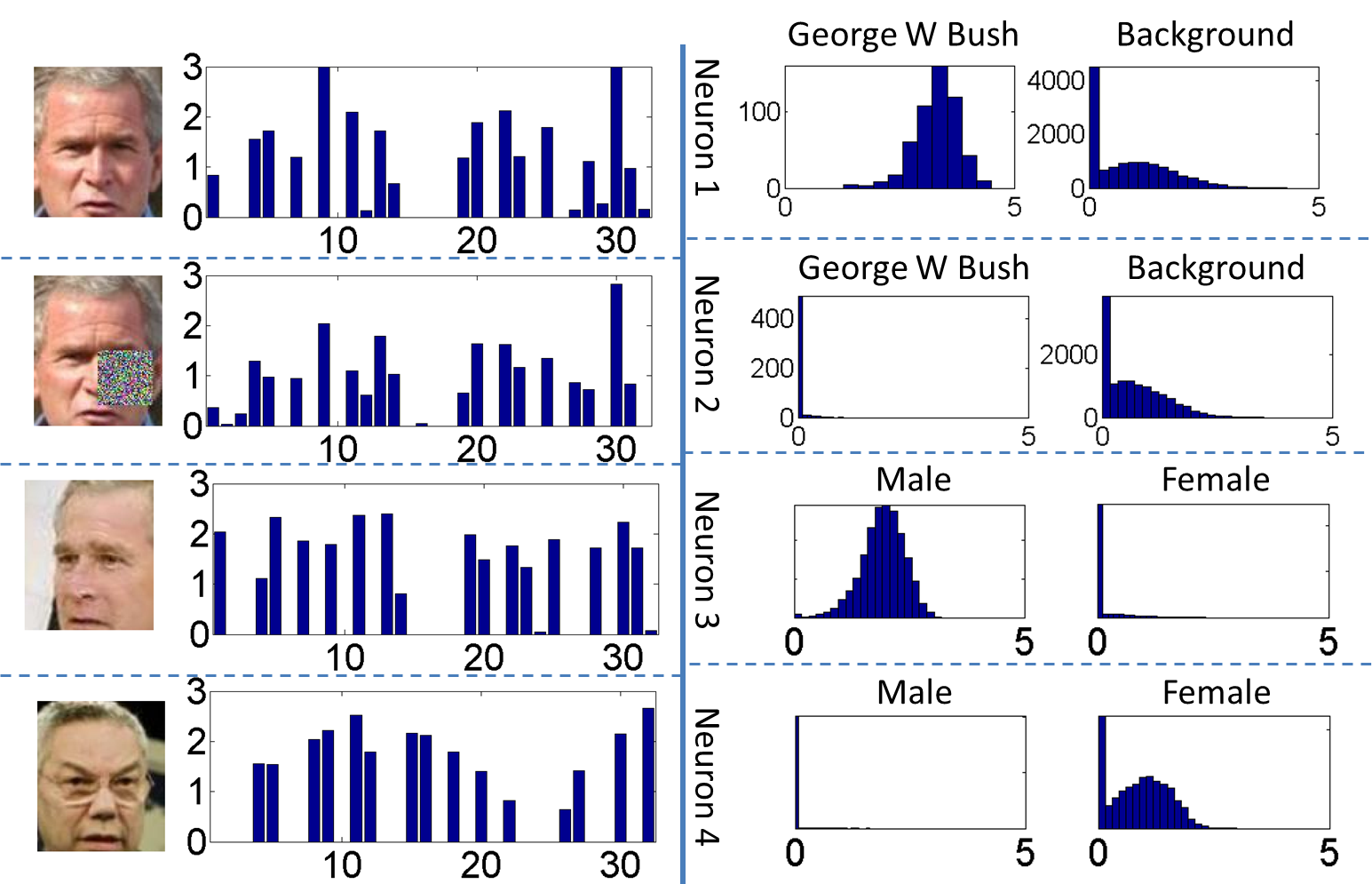}
\end{center}
\vspace{-0.15in}
\caption{Left: neural responses of DeepID2+ on images of Bush and Powell.  The second face is partially occluded. There are 512 neurons in the top hidden layer of DeepID2+. We subsample 32 for illustration. Right: a few neurons are selected to show their activation histograms over all the LFW face images (as background), all the images belonging to Bush, all the images with attribute Male, and all the images with attribute Female. A neuron is generally activated on about half of the face images. But it may constantly have activations (or no activation) for all the images belonging to a particular person or attribute.  In this sense, neurons are sparse, and selective to identities and attributes. }
\label{fig:demo}
\vspace{-0.1in}
\end{figure}

Our study is based on a high-performance deep convolutional neural network (deep ConvNet \cite{lecun1989,lecun1998}), referred to as DeepID2+, proposed in this paper. It is improved upon the  state-of-the-art DeepID2 net \cite{sun2014b} by increasing the dimension of hidden representations and adding  supervision to early convolutional layers. The best single DeepID2+ net (taking both the original and horizontally flipped face images as input) achieves $98.70\%$ verification accuracy on LFW (vs. $96.72\%$ by DeepID2). Combining $25$ DeepID2+ nets sets new state-of-the-art on multiple benchmarks:  $99.47\%$ on LFW for face verification (vs. $99.15\%$ by DeepID2 \cite{sun2014b}),  $95.0\%$ and $80.7\%$ on LFW for closed- and open-set face identification, respectively (vs. $82.5\%$ and $61.9\%$ by Web-Scale Training (WST) \cite{taigman2014b}), and $93.2\%$ on YouTubeFaces \cite{wolf2011} for face verification (vs. $91.4\%$ by DeepFace \cite{taigman2014a}).

With the state-of-the-art deep ConvNets and through extensive empirical evaluation, we investigate three properties of  neural activations crucial for  the high performance: sparsity, selectiveness, and robustness. They are naturally owned by deepID2+ after large scale training on face data, and we did NOT enforce any extra regularization to the model and training process to achieve them. Therefore, these results are valuable for understanding the intrinsic properties of deep  networks.

It is observed that the neural activations of DeepID2+ are moderately sparse. As examples shown in Fig. \ref{fig:demo}, for an input face image, around half of the neurons in the top hidden layer are activated. On the other hand, each neuron is activated on roughly half of the face images. Such sparsity distributions can maximize the discriminative power of  the deep net as well as the distance between images. Different identities have different subsets of neurons activated. Two images of the same identity have similar activation patterns. This motivates us to binarize the neural responses in the top hidden layer and  use the binary code for  recognition. Its result is surprisingly good. Its verification accuracy on LFW only slightly drops by $1\%$ or less. It has significant impact on large-scale face search since huge storage and computation time is saved. This also implies that binary activation patterns are more important than activation magnitudes in deep neural networks.

Related to sparseness, it is also observed that neurons in higher layers are highly selective to identities and identity-related attributes. When an identity (who can be outside the training data) or attribute is presented, we can identify a subset of neurons which are constantly excited and also can find another subset of neurons which are constantly inhibited. A neuron from any of these two subsets has strong indication on the existence/non-existence of this identity or attribute, and our experiment show that the single neuron alone has high recognition accuracy for a particular identity or attribute. In other words, neural activations have sparsity on identities and attributes, as examples shown in Fig. \ref{fig:demo}. Although  DeepID2+  is not taught to distinguish attributes during training, it has implicitly learned such high-level concepts. Directly employing the face representation learned by DeepID2+ leads to much higher classification accuracy on identity-related attributes than widely used handcrafted features such as high-dimensional LBP \cite{chen2013,cao2013}.

Our empirical study shows that neurons in higher layers are much more robust to image corruption in face recognition than handcrafted features such as high-dimensional LBP or neurons in lower layers. As an example shown in Fig. \ref{fig:demo}, when a face image is partially occluded, its binary activation patterns remain stable, although the magnitudes could change. We conjecture the reason might be that neurons in higher layers capture global features and are less sensitive to local variations. DeepID2+ is trained by natural web face images and no artificial occlusion patterns were added to the training set.

\section{Related work}

Only very recently, deep learning  achieved great success on face recognition \cite{zhu2013,taigman2014a,sun2014a,sun2014b,taigman2014b} and significantly outperformed systems using low level features \cite{kumar2009,taigman2009,guillaumin2009,yin2011,huang2011,chen2012,berg2012,simonyan2013,chen2013,cao2013}. There are two notable breakthroughs. The first is large-scale face identification with deep neural networks \cite{taigman2014a,sun2014a,taigman2014b}. By classifying face images into thousands or even millions of identities, the last hidden layer  forms features highly discriminative to identities. The second is supervising deep neural networks with both face identification and verification tasks \cite{sun2014b}. The verification task minimizes the distance between features of the same identity, and decreases  intra-personal variations \cite{sun2014b}. By combining features learned  from many face regions, \cite{sun2014b} achieved the current state-of-the-art ($99.15\%$) of face verification on LFW.

Attribute learning  is  an active  topic \cite{farhadi2009,berg2010,parikh2011,bourdev2011,luo2013,zhang2014}. There have been works on first learning attribute classifiers and using attribute predictions for face recognition \cite{kumar2009,chung2012}. What we have tried in this paper is the inverse, by first predicting the identities, and then using the learned identity-related features to predict attributes.

Sparse representation-based classification \cite{wright2009, yang2010,liu2010,zhang2011a,elhamifar2011,jia2012} was extensively studied for face recognition with occlusions. Tang \etal \cite{tang2012} proposed Robust Boltzmann Machine to distinguish corrupted pixels and learn latent representations. These  methods designed components explicitly handling occlusions, while we show that features learned by DeepID2+ have implicitly encoded  invariance to occlusions. This is naturally achieved without adding regulation to  models or artificial occlusion patterns to  training data.

\section{DeepID2+ nets}

Our DeepID2+ nets are inherited from DeepID2 nets \cite{sun2014b}, which have four convolutional layers, with $20$, $40$, $60$, and $80$  feature maps, followed by a $160$-dimensional feature layer fully-connected to both the third and fourth convolutional layers. The $160$-dimensional feature layer (DeepID2 feature layer) is supervised by both face identification and verification tasks. Given a pair of training  images, it  obtains two DeepID2 feature vectors ($f_i$ and $f_j$) by forward-propagating the two images through the DeepID2 net. Then each DeepID2 feature vector is classified as one of $8192$ identities in the training set, and the classification (identification) error is generated. The verification error is $\frac{1}{2}\left\|f_i-f_j\right\|_2^2$ if $f_i$ and $f_j$ are from the same identity, or $\frac{1}{2}\max\left(0, m - \left\|f_i-f_j\right\|_2\right)^2$ otherwise. It was shown that combining identification and verification supervisory signals helps to learn features more effectively \cite{sun2014b}.

\begin{figure}[t]
\begin{center}
\includegraphics[width=0.8\linewidth]{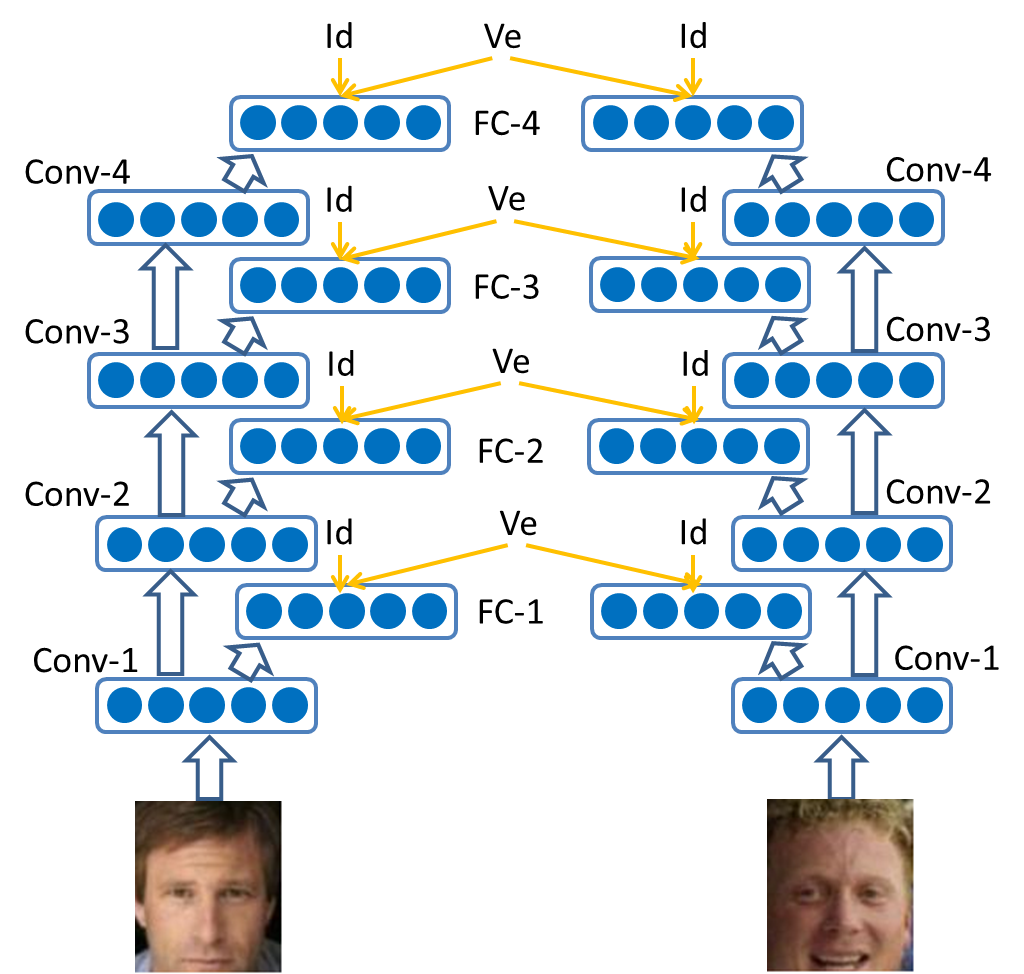}
\end{center}
\vspace{-0.15in}
\caption{DeepID2+ net and  supervisory signals. Conv-$n$ deneotes the $n$-th convolutional layer (with max-pooling). FC-$n$ denotes the $n$-th fully connected layer. Id and Ve denote the identification and verification supervisory signals. Blue arrows denote forward-propagation. Yellow arrows denote supervisory signals. Nets in the left and right are the same DeepID2+ net with different input faces.}
\label{fig:cnn}
\vspace{-0.1in}
\end{figure}

Compared to DeepID2, DeepID2+ makes three improvements. First, it is larger with $128$ feature maps in each of the four convolutional layers. The final feature representation is increased from $160$ to $512$ dimensions. Second, our training data is  enlarged by merging the CelebFaces+ dataset\cite{sun2014a}, the WDRef dataset \cite{chen2012}, and some newly collected identities exclusive from LFW. The larger DeepID2+ net is trained with around $290,000$ face images from $12,000$ identities compared to $160,000$ images from $8,000$ identities used to train the DeepID2 net. Third, in the DeepID2 net, supervisory signals are only added to one fully-connected layer connected to the third and fourth convolutional layers, while the lower convolutional layers can only get supervision with gradients back-propagated from higher layers. We enhance the supervision by connecting a $512$-dimensional fully-connected layer to each of the four convolutional layers (after max-pooling), denoted as FC-$n$ for $n=1,2,3,4$, and supervise these four fully-connected layers with the identification-verification supervisory signals \cite{sun2014b} simultaneously as shown in Fig. \ref{fig:cnn}. In this way, supervisory signals become "closer" to the early convolutional layers and are more effective.

\section{High-performance of DeepID2+ nets}
\label{sec:performance}

\begin{figure}[t]
\begin{center}
\includegraphics[width=1.0\linewidth]{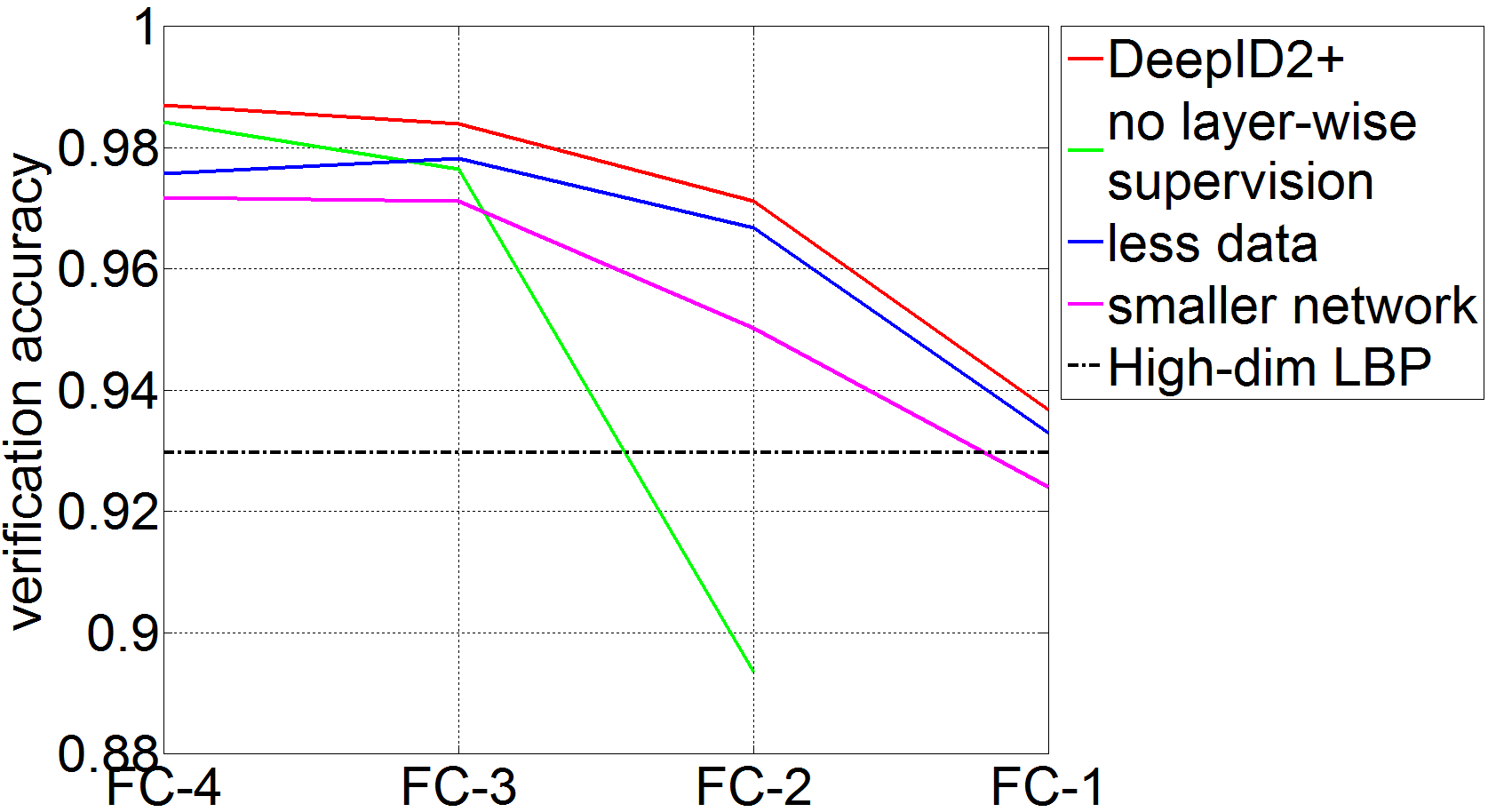}
\end{center}
\vspace{-0.15in}
\caption{Comparison of DeepID2+ net and those with no layer-wise supervision, less training data, and fewer feature maps, respectively.}
\label{fig:self_comparison}
\vspace{-0.1in}
\end{figure}

To verify the improvements, we learn the Joint Bayesian model \cite{chen2012} for face verification based on each of the four $512$-dimensional feature vectors (neural activations) FC-n for $n=1,2,3,4$ in the DeepID2+ net. Joint Bayesian is trained on $2000$ people in our training set (exclusive from people in LFW) which are not used for training DeepID2+ net, and tested on the $6,000$ given face pairs in LFW for face verification. These $2000$ identities also serve as a validating set when training the DeepID2+ net to determine learning rates and training iterations. The DeepID2+ net (proposed) is compared to three nets with one of the three improvements removed, respectively, as shown in Fig. \ref{fig:self_comparison}. For the net with no layer-wise supervision, only the gradients of FC-4 is back-propagated to the convolutional layers. For the net with less training data, only the training data from CelebFaces+ is used. For the smaller network, the numbers of feature maps in  convolutional layers are the same as those in DeepID2 and 160-dimensional features are used for FC-n for $n=1,2,3,4$. All the networks compared are learned on a fixed region covering the entire face. We can clearly see the improvements of the three aspects from Fig. \ref{fig:self_comparison}.

\begin{figure}[t]
\begin{center}
\includegraphics[width=1.0\linewidth]{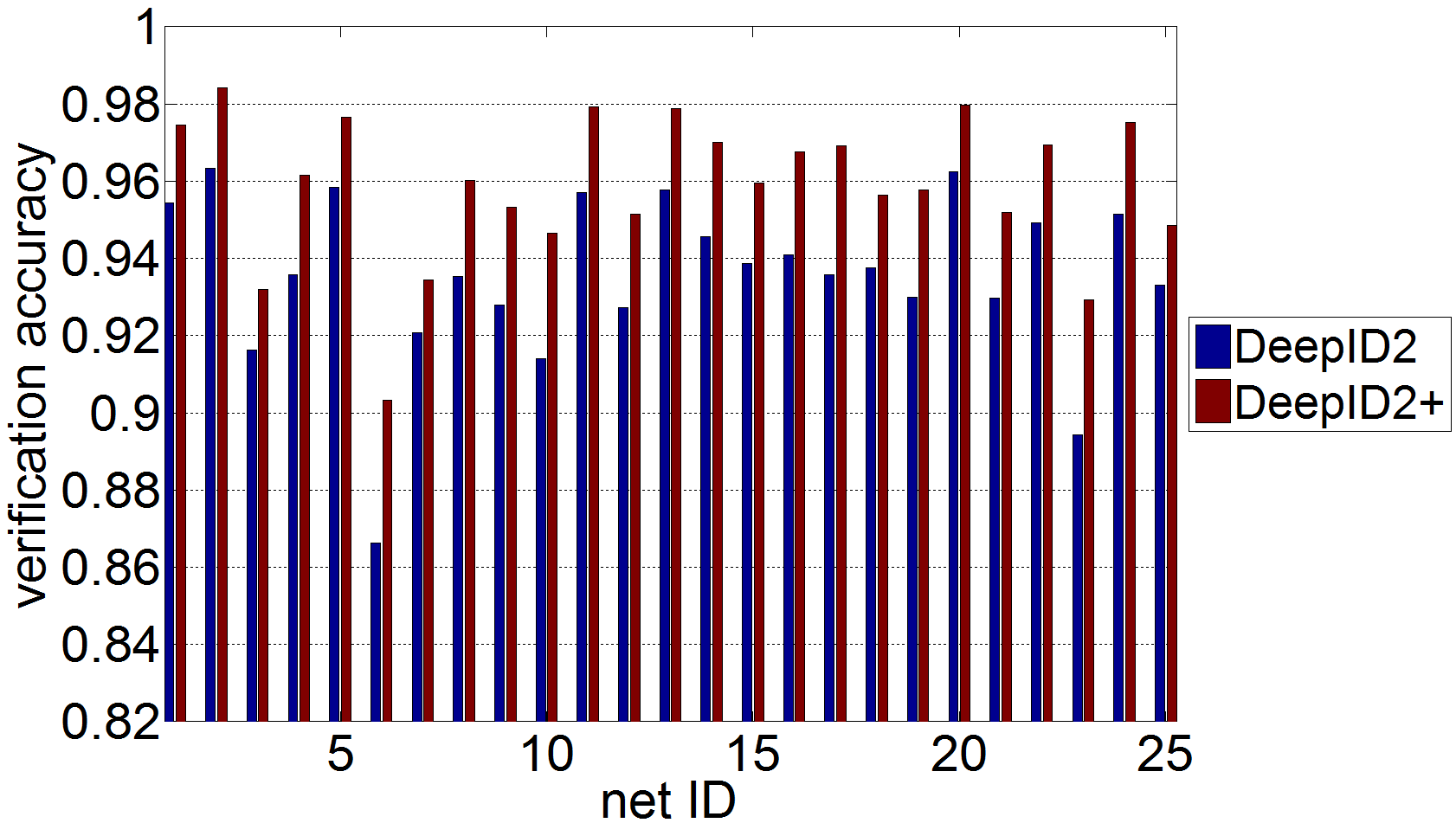}
\end{center}
\vspace{-0.15in}
\caption{Comparison of face verification accuracies on LFW with ConvNets trained on $25$ face regions given in DeepID2 \cite{sun2014b}}
\label{fig:net25_noflip}
\vspace{-0.1in}
\end{figure}

\begin{figure}[!h]
\begin{center}
\includegraphics[width = 0.8\linewidth]{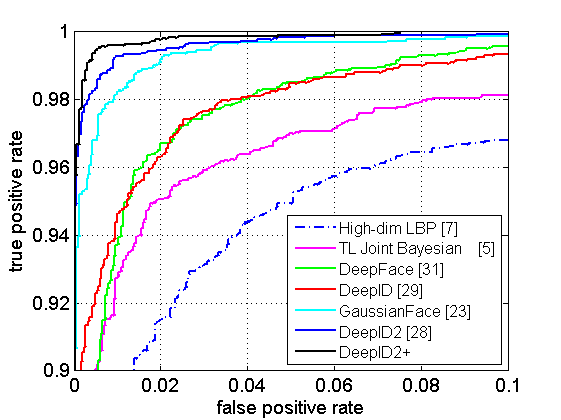}
\end{center}
\vspace{-0.15in}
\caption{ROC  of face verification on LFW. Best viewed in color.}
\label{fig:lfw}
\vspace{-0.1in}
\end{figure}

\begin{figure}[!h]
\begin{center}
\includegraphics[width = 0.8\linewidth]{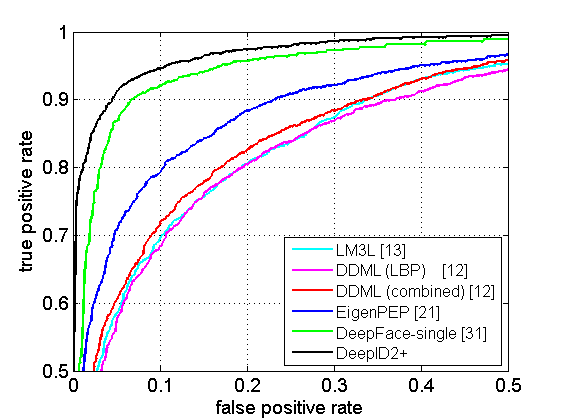}
\end{center}
\vspace{-0.15in}
\caption{ROC  of face verification on YouTube Faces. Best viewed in color.}
\label{fig:youtube}
\vspace{-0.05in}
\end{figure}

To compare with DeepID2 nets \cite{sun2014b}, we train $25$ DeepID2+ nets taking the same $25$ face regions selected by DeepID2 as shown in Fig. 2 in \cite{sun2014b}. Features in the FC-$4$ layer of DeepID2+ are extracted to train Joint Bayesian for face verification (features are extracted on either the original or the horizontally flipped face regions as shown in Fig. 2 in \cite{sun2014b}). The comparison between the $25$ deep ConvNets on the LFW face verification task is shown in Fig. \ref{fig:net25_noflip}. DeepID2+ nets improve approximately $2\%$ accuracy on average over DeepID2. When combining FC-$4$ layer features extracted from all the $25$ face regions and their horizontally flipped counterparts with the $25$ DeepID2+ nets, respectively, we achieve $\bm{99.47\%}$ and $\bm{93.2\%}$ face verification accuracies on LFW and YouTube Faces datasets, respectively. Tab. \ref{tab:lfw} and Tab. \ref{tab:youtube} are accuracy comparisons with the previous best results on the two datasets. Fig. \ref{fig:lfw} and Fig. \ref{fig:youtube} are the  ROC  comparisons. Our DeepID2+ nets outperform all previous results on both datasets. There are a few wrongly labeled test face pairs in LFW and YouTubeFaces. After correction, our face verification accuracy  increases  to $99.52\%$ on LFW and $93.8\%$ on YouTubeFaces.

Face identification is a more challenging task to evaluate high-performance face recognition systems \cite{taigman2014b}. Therefore we further evaluate the $25$ DeepID2+ nets on the closed- and open-set face identification tasks on LFW, following the protocol  in \cite{best-rowden2014}. The closed-set identification reports the Rank-$1$ identification accuracy while the open-set identification  reports the Rank-$1$ Detection and Identification rate (DIR) at a $1\%$ False Alarm Rate (FAR). The comparison results are shown in Tab. \ref{tab:lfw_id}. Our results significantly outperform the previous best \cite{taigman2014b} with $\bm{95.0\%}$ and $\bm{80.7\%}$ closed and open-set identification accuracies, respectively.

\begin{table}[t]
\caption{Face verification on LFW.}
\label{tab:lfw}
\vspace{-0.1in}
\begin{center}
\begin{tabular}{p{100pt}|p{100pt}}
\toprule
method & accuracy (\%) \\
\midrule
High-dim LBP \cite{chen2013} & $95.17\pm1.13$ \\
TL Joint Bayesian \cite{cao2013} & $96.33\pm1.08$ \\
DeepFace \cite{taigman2014a} & $97.35\pm0.25$ \\
DeepID \cite{sun2014a} & $97.45\pm0.26$ \\
GaussianFace \cite{lu2014} & $98.52\pm0.66$ \\
DeepID2 \cite{sun2014b} & $99.15\pm0.13$ \\
DeepID2+ & $\bm{99.47\pm0.12}$ \\
\bottomrule
\end{tabular}
\end{center}
\vspace{-0.1in}
\end{table}

\begin{table}[t]
\caption{Face verification on YouTube Faces.}
\label{tab:youtube}
\vspace{-0.1in}
\begin{center}
\begin{tabular}{p{100pt}|p{100pt}}
\toprule
method & accuracy (\%) \\
\midrule
LM3L \cite{hu2014b} & $81.3\pm1.2$ \\
DDML (LBP) \cite{hu2014a} & $81.3\pm1.6$ \\
DDML (combined) \cite{hu2014a} & $82.3\pm1.5$ \\
EigenPEP \cite{li2014} & $84.8\pm1.4$ \\
DeepFace-single \cite{taigman2014a} & $91.4\pm1.1$ \\
DeepID2+ & $\bm{93.2\pm0.2}$ \\
\bottomrule
\end{tabular}
\end{center}
\vspace{-0.1in}
\end{table}

\begin{table}[t]
\caption{Closed- and open-set identification tasks on LFW. }
\vspace{-0.1in}
\label{tab:lfw_id}
\begin{center}
\begin{tabular}{p{80pt}|p{50pt}|p{50pt}}
\toprule
method & Rank-1 (\%) & DIR $@$ $1\%$ FAR ($\%$) \\
\midrule
COTS-s1 \cite{best-rowden2014} & $56.7$ & $25$ \\
COTS-s1+s4 \cite{best-rowden2014} & $66.5$ & $35$ \\
DeepFace \cite{taigman2014a} & $64.9$ & $44.5$ \\
WST Fusion \cite{taigman2014b} & $82.5$ & $61.9$ \\
DeepID2+ & $\bm{95.0}$ & $\bm{80.7}$ \\
\bottomrule
\end{tabular}
\end{center}
\vspace{-0.1in}
\end{table}

\section{Moderate sparsity of neural activations}

\begin{figure}[!h]
\begin{center}
\includegraphics[width = 0.9\linewidth]{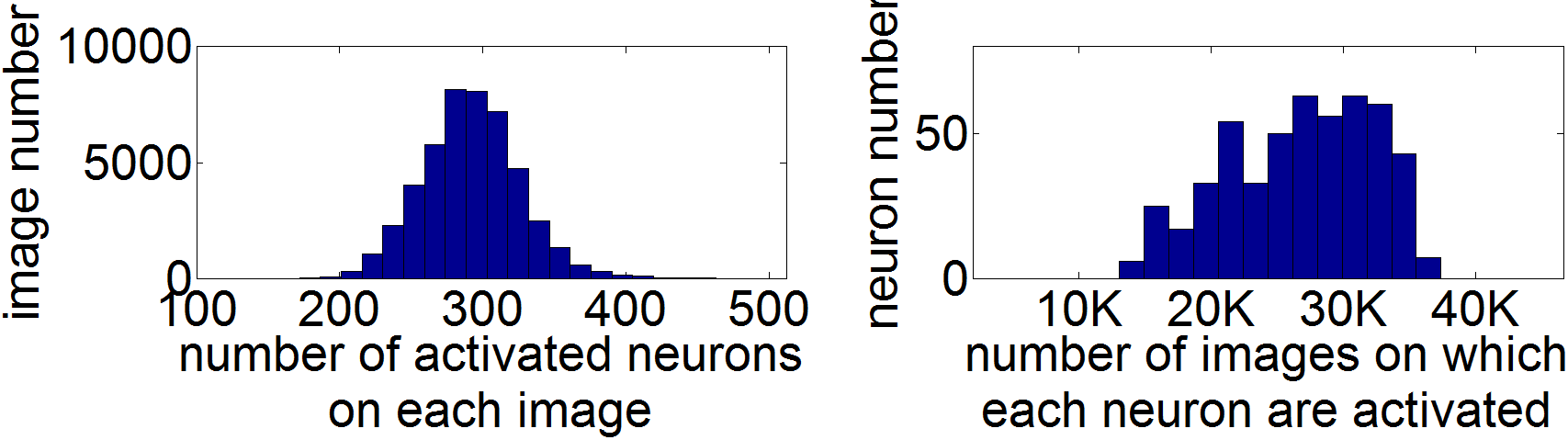}
\end{center}
\vspace{-0.15in}
\caption{Left: the histogram of the number of activated neurons for each of the validating images. Right: the histogram of the number of images on which each neuron is activated.}
\label{fig:sparsity}
\vspace{-0.05in}
\end{figure}

Neural activations are moderately sparse in both the sense that for each image, there are approximately half of the neurons which are activated (with positive activation values) on it, and for each neuron, there are approximately half of the images on which it is activated. The moderate sparsity on images makes faces of different identities maximally distinguishable, while the moderate sparsity on neurons makes them to have maximum discrimination abilities. We verify this by calculating the histogram of the activated neural numbers on each of the $46,594$ images in our validating dataset (Fig. \ref{fig:sparsity} left), and the histogram of the number of images on which each neuron are activated (Fig. \ref{fig:sparsity} right). The evaluation is based on the FC-$4$ layer neurons in a single DeepID2+ net taking the entire face region as input. Compared to all $512$ neurons in the FC-$4$ layer, the mean and standard deviation of the number of activated neurons on images is $292\pm34$, while compared to all $46,594$ validating images, the mean and standard deviation of the number of images on which each neuron are activated is $26,565\pm5754$, both of which are approximated centered at half of all neurons/images.

We further verify that the activation patterns, \ie, whether neurons are activated, are more important than precise activation values. We convert neural activations to binary code by thresholding and compare its face verification ability on LFW to that of the original representation. As shown in Tab. \ref{tab:sparsity}, the binary representation, when coupled with Joint Bayesian, sacrifices $1\%$ or less accuracies ($97.67\%$ and $99.12\%$ with a single net or combining $25$ nets, respectively). More interestingly, the binary code can still achieve $96.45\%$ and $97.47\%$ accuracy with a single net or combining $25$ nets, respectively, even by directly calculating the Hamming distances. This shows that the state of excitation or inhibition of neurons already contains the majority of discriminative information. Binary code is economic for storage and fast for image search. We believe this would be an interesting direction of future work.

\begin{table}[t]
\caption{Comparison of the original DeepID2+ features and its binarized representation for face verification on LFW. The first two rows of results are accuracies of the original (real-valued) FC-$4$ layer representation of a single net (real single) and of the $25$ nets (real comb.), respectively, with Joint Bayesian as the similarity metrics. The last two rows of results are accuracies of the corresponding binary representations, with Joint Bayesian or Hamming distance as the similarity metrics, respectively.}
\vspace{-0.1in}
\label{tab:sparsity}
\begin{center}
\begin{tabular}{p{60pt}|p{60pt}|p{60pt}}
\toprule
 & Joint Bayesian ($\%$) & Hamming distance ($\%$) \\
\midrule
real single & $98.70$ & N/A \\
real comb. & $99.47$ & N/A \\
binary single & $97.67$ & $96.45$ \\
binary comb. & $99.12$ & $97.47$ \\
\bottomrule
\end{tabular}
\end{center}
\vspace{-0.1in}
\end{table}

\section{Selectiveness on identities and attributes}
\label{sec:selectiveness}

\subsection{Discriminative power of neurons}
\begin{figure}[t]
\begin{center}
\includegraphics[width=1.0\linewidth]{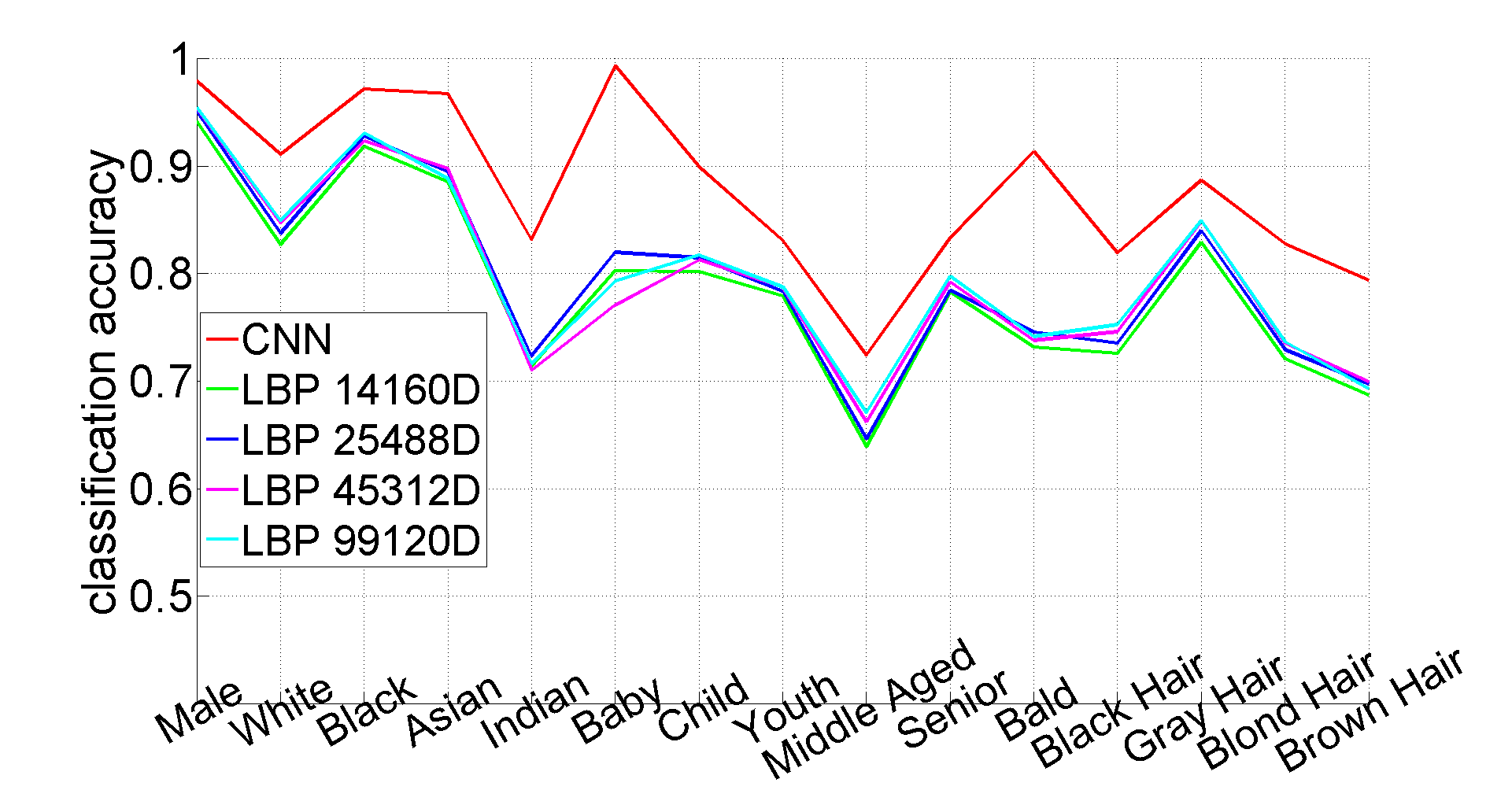}
\end{center}
\vspace{-0.15in}
\caption{Accuracy comparison between DeepID2+ and LBP features for attribute classification on LFW.}
\label{fig:attribute_cv}
\vspace{-0.1in}
\end{figure}

\begin{figure}[t]
\begin{center}
\includegraphics[width=0.8\linewidth]{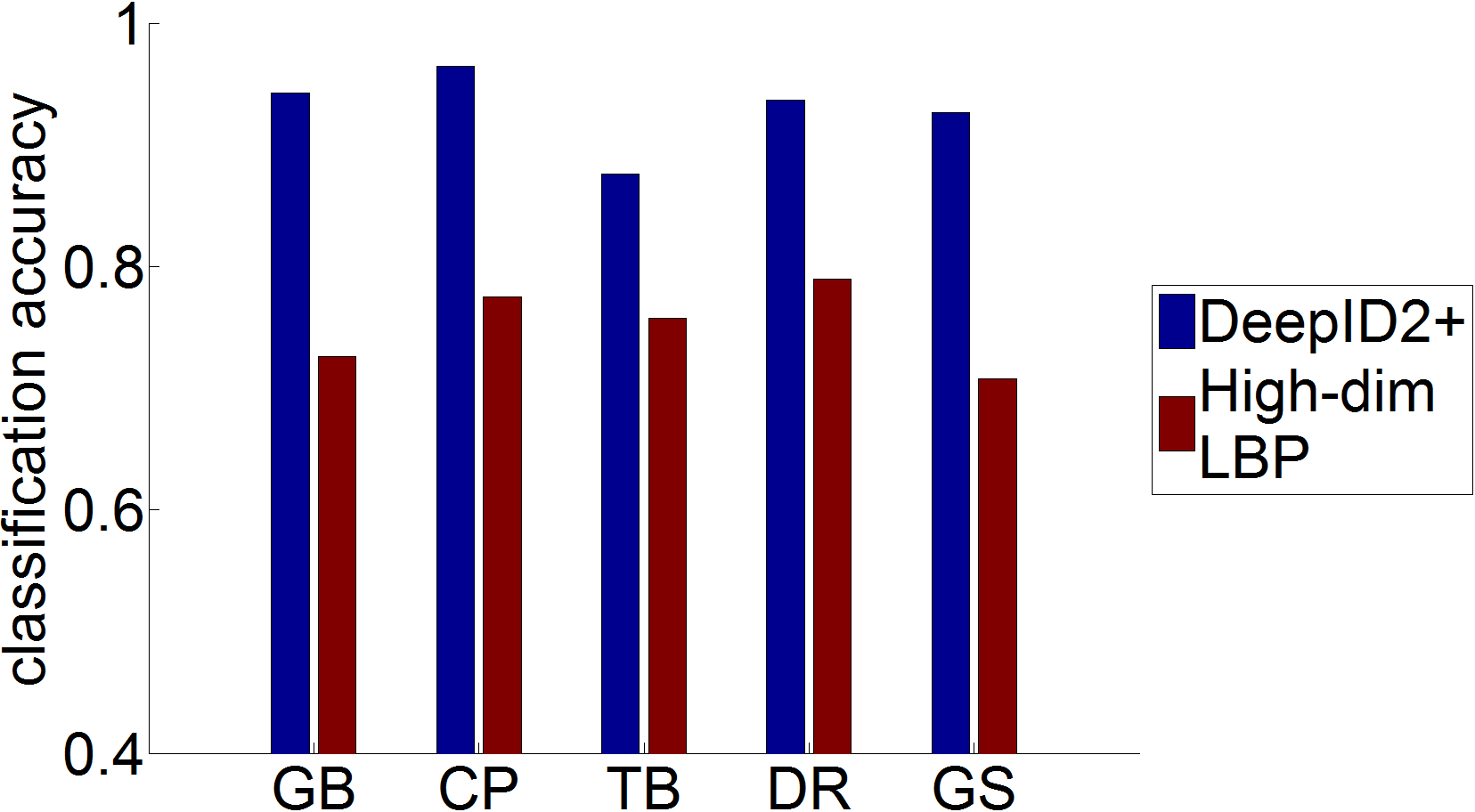}
\end{center}
\vspace{-0.15in}
\caption{Identity classification accuracy on LFW with one single DeepID2+ or LBP feature. Initials of identity names are used.}
\label{fig:identity_sel}
\vspace{-0.1in}
\end{figure}

\begin{figure}[t]
\begin{center}
\includegraphics[width=0.8\linewidth]{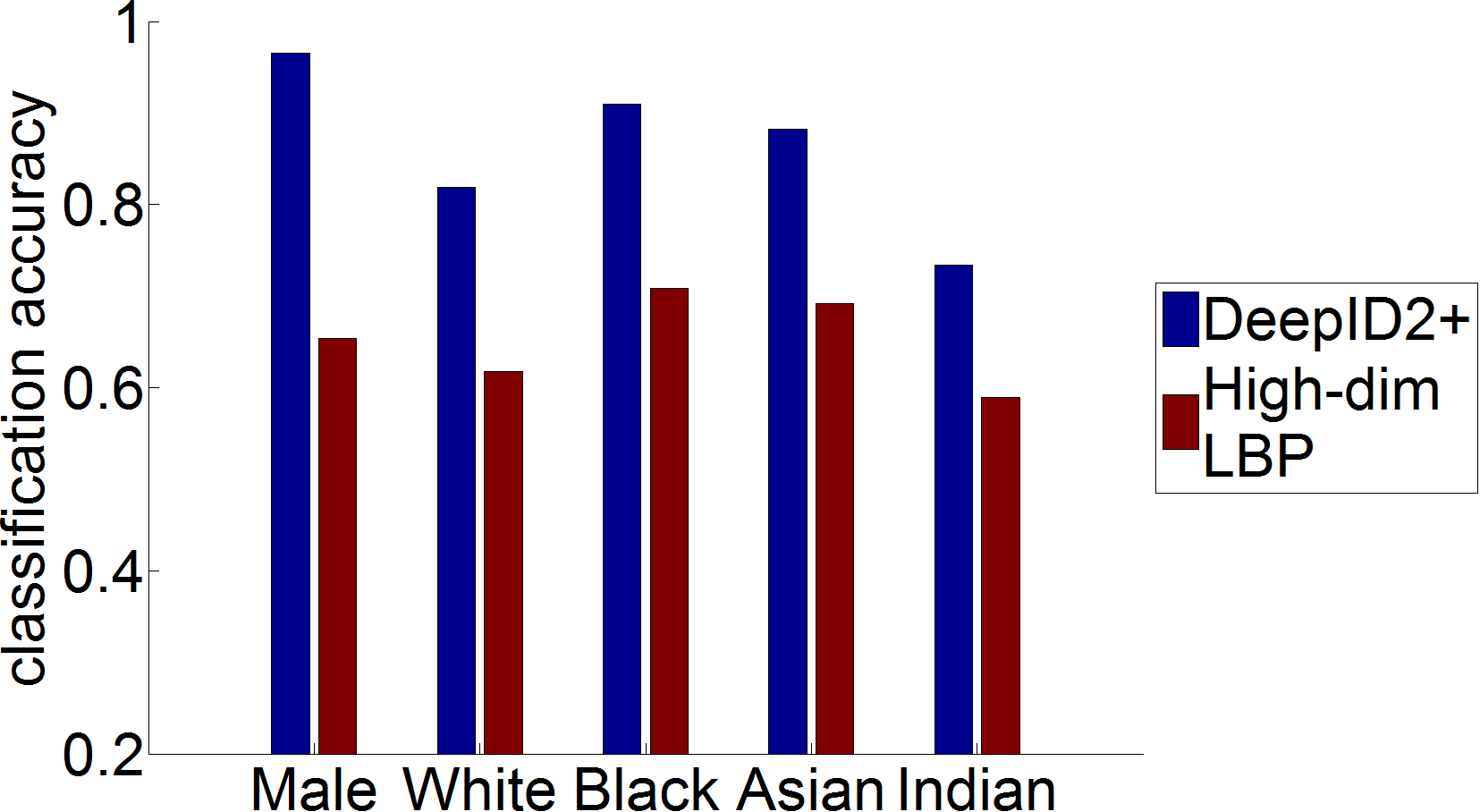}
\end{center}
\vspace{-0.15in}
\caption{Attribute classification accuracy on LFW with one single DeepID2+ or LBP feature.}
\label{fig:attribute_sel}
\vspace{-0.1in}
\end{figure}

We test DeepID2+ features for two binary classification tasks. The first is to classify the face images of one person against those of all the other people or the background. The second is to classify a face image as having an attribute or not. DeepID2+ features are taken from the FC-$4$ layer of a single DeepID2+ net on the entire face region and its horizontally flipped counterpart, respectively. The experiments are conducted on LFW with people unseen by the DeepID2+ net during training.  LFW is randomly split into two subsets and the cross-validation accuracies are reported. The accuracies are normalized \wrt the image numbers in the positive and negative classes. We also compare to the high-dimensional LBP features \cite{chen2013,cao2013} with various feature dimensions. As shown in Fig. \ref{fig:attribute_cv}, DeepID2+ features significantly outperform LBP in attribute classification (it is not surprising that DeepID2+ has good identity classification result). Fig. \ref{fig:identity_sel} and Fig. \ref{fig:attribute_sel} show identity and attribute classification accuracies with only one best feature selected. Different best features are selected for different identities (attributes). With a single feature (neuron), DeepID2+ reaches approximately $97\%$ for some identity and attribute. This is the evidence that DeepID2+ features are identity and attribute selective. Apparently LBP does not have it.

\subsection{Excitatory and inhibitory neurons}
\begin{figure}[tb!]

\begin{subfigure}{1.0\linewidth}
\includegraphics[width = 1.0\linewidth]{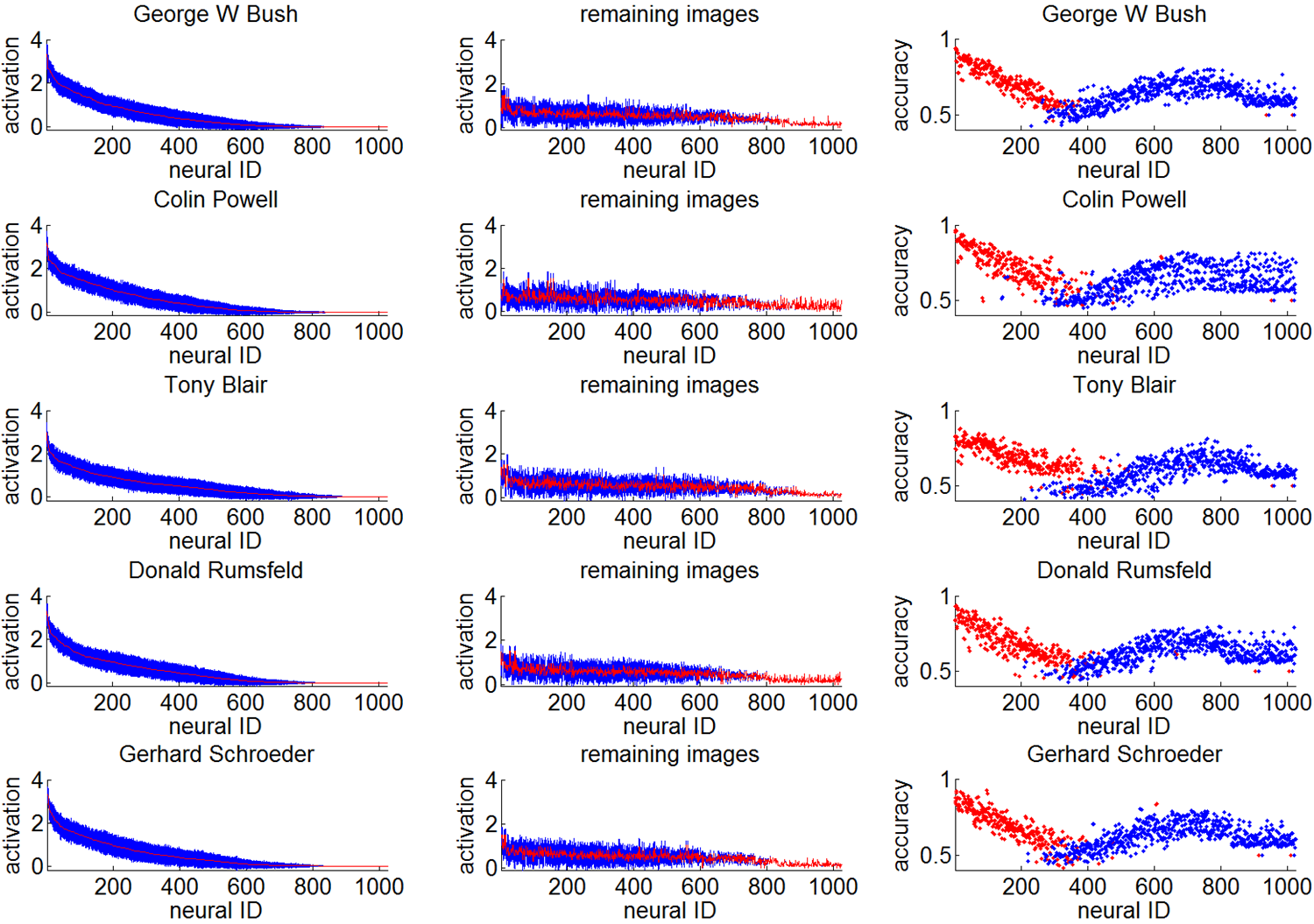}
\vspace{-0.2in}
\caption{DeepID2+ neural activation distributions and per-neuron classification accuracies.}
\label{fig:actcnn}
\vspace{0.05in}
\end{subfigure}

\begin{subfigure}{1.0\linewidth}
\includegraphics[width = 1.0\linewidth]{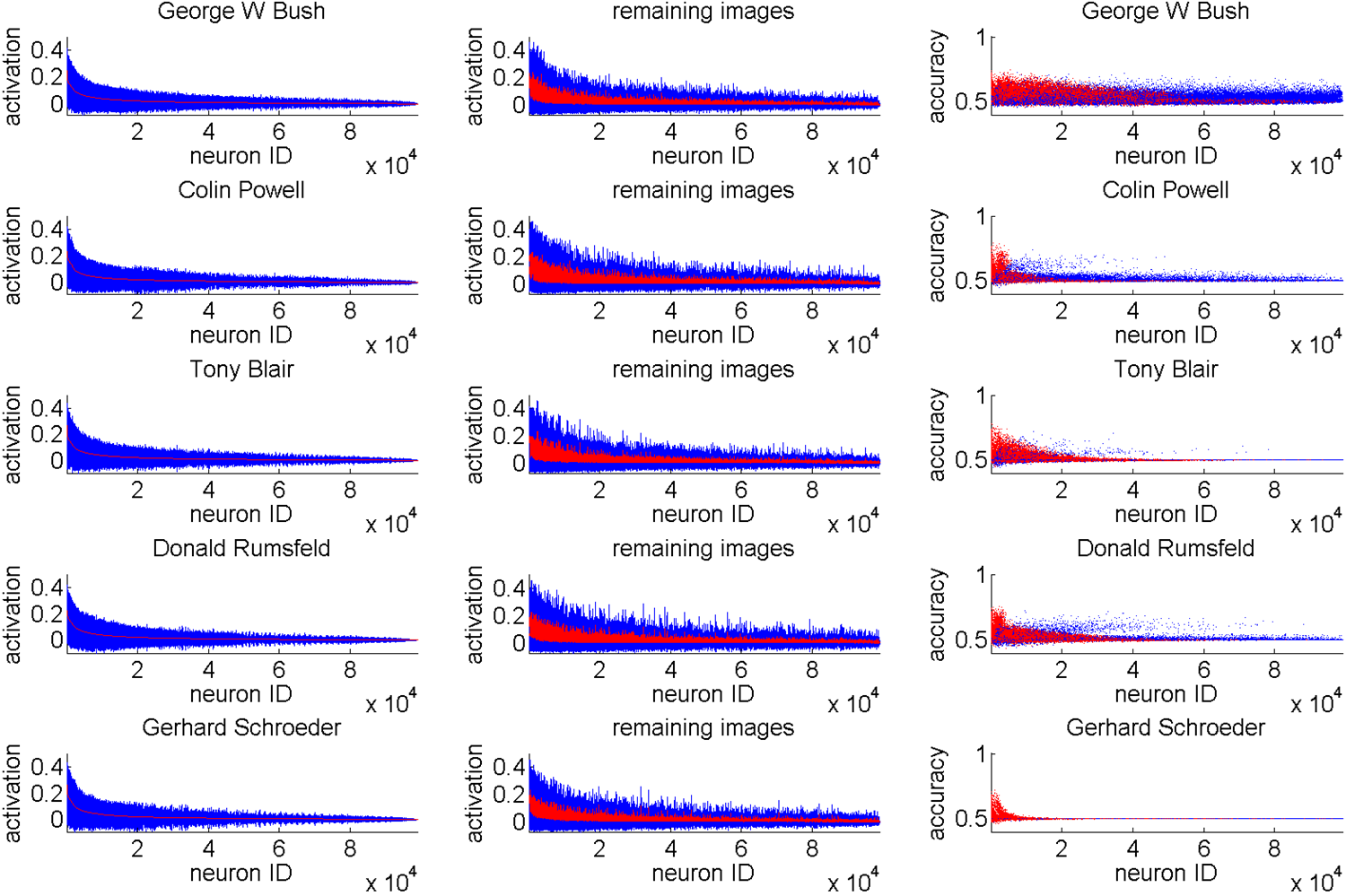}
\vspace{-0.2in}
\caption{LBP feature activation distributions and per-feature classification accuracies.}
\label{fig:actlbp}
\vspace{0.0in}
\end{subfigure}

\vspace{-0.05in}
\caption{Comparison of distributions of DeepID2+ neural and LBP feature activations and per-neuron (feature) classification accuracies for the first five people with the most face images in LFW. Left column: mean and standard deviations of neural (feature) activations on images belonging to a single identity. Mean is represented by a red line while standard deviations are represented by vertical segments between (mean $-$ standard deviation) and (mean $+$ standard deviation). Neurons (features) are sorted by their mean activations on the given identity. Middle column: mean and standard deviations of neural (feature) activations on the remaining images. Neural (feature) orders are the same as those in the left column. Right column: per-neuron (feature) classification accuracies on the given identity. Neural (feature) orders are the same as those in the left and middle columns. Neurons (features) activated and inhibited for a given identity are marked as red and blue dots, respectively.}
\vspace{-0.05in}
\label{fig:act}
\end{figure}

We find that the discrimination to identities and facial attributes are due to neurons' excitation and inhibition patterns on certain identities or attributes. For example, a neuron may be excited when it sees George Bush while becoming inhibitive when it sees Colin Powell, or a neuron may be excited for western people while being inhibitive for Asian. Fig. \ref{fig:actcnn} compares the mean and standard deviation of DeepID2+ neural activations over images belonging to a particular single identity (left column) and over the remaining images (middle column), as well as showing the per-neuron classification accuracies of distinguishing each given identity from the remaining images (right column). The top five identities with the most face images in LFW are evaluated (the other identities have similar results). Neural orders are sorted by the mean neural activations on the evaluated identity for figures in all the three columns. For each given identity there are neurons strongly excited (\eg, those with neural ID smaller than $200$) or inhibited (\eg, those with neural ID larger than $600$). For the excited neurons, their activations are distributed in higher values, while  other images have significantly lower mean values on these neurons. Therefore, the excitatory neurons can easily distinguish an identity from others, which is verified by their high classification accuracies shown by the red dots with small neural IDs in figures in the right column.

For neurons ranked in the middle (\eg, those with neural ID around $400$), their activation distributions on the given identity are largely overlapped with those on other identities. They have weak discrimination abilities for the given identity, verified by the low accuracies of the red and blue dots near the junction of the two colors. The excitation or inhibition state of these neurons has much uncertainty.

When mean activations further decrease (\eg, neural ID above $600$), the neurons demonstrate inhibitory properties, and are seldom activated for the given identity compared to others. These inhibitory neurons also have discrimination abilities with relatively high classification accuracies.

However, similar phenomena cannot be found on LBP features as shown in Fig. \ref{fig:actlbp}. The range of LBP features on given identities and the remaining images are overlapped for all features. Compared to DeepID2+ neural activations, LBP features have much lower classification accuracies, the majority of which are accumulated on the $50\%$ random guess line

\begin{figure}[tb!]
\centering

\begin{subfigure}{0.98\linewidth}
\includegraphics[width = 0.98\linewidth]{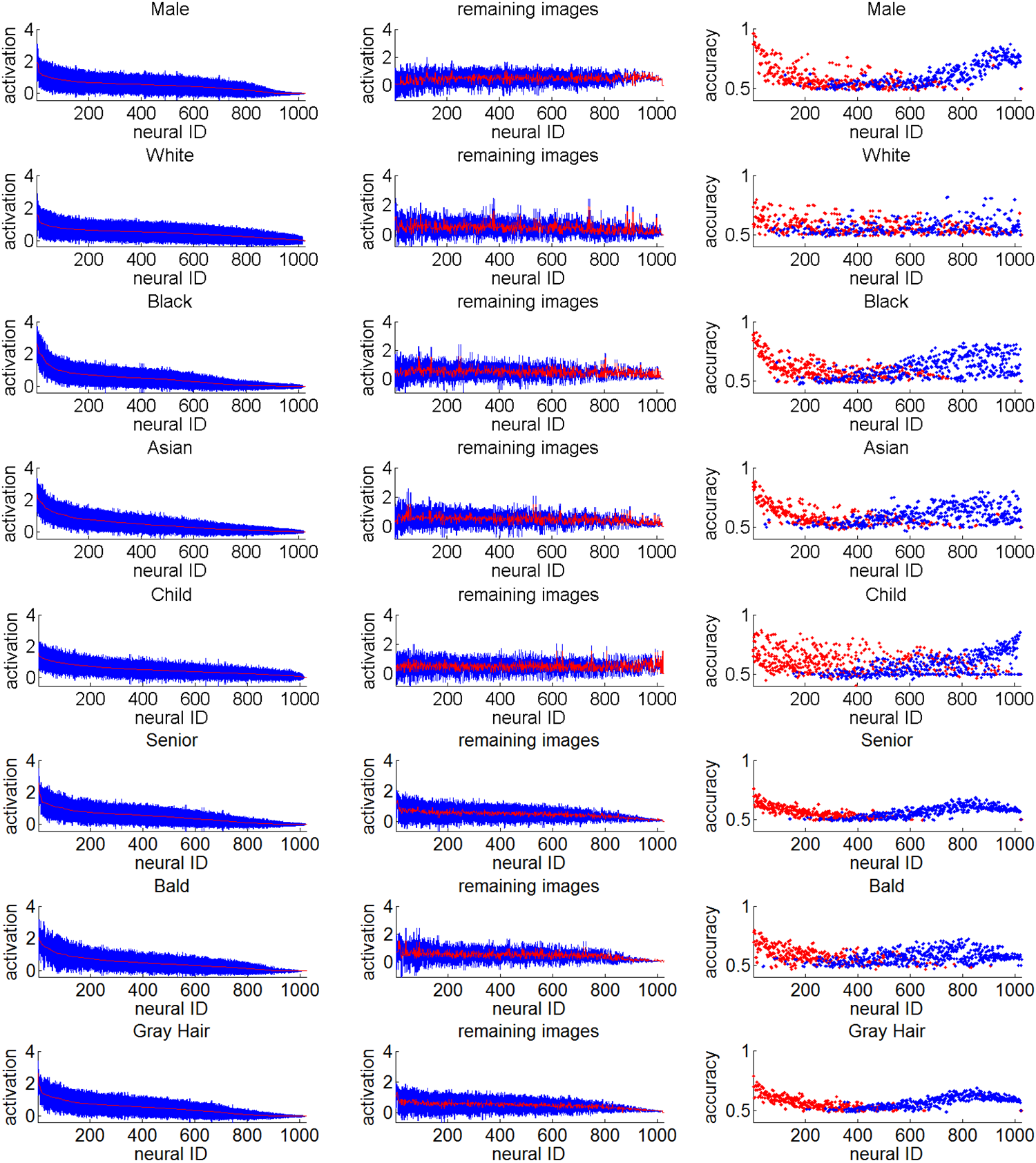}
\vspace{-0.02in}
\caption{DeepID2+ neural activation distributions and per-neuron classification accuracies.}
\label{fig:attcnn}
\vspace{0.0in}
\end{subfigure}

\begin{subfigure}{0.98\linewidth}
\includegraphics[width = 0.98\linewidth]{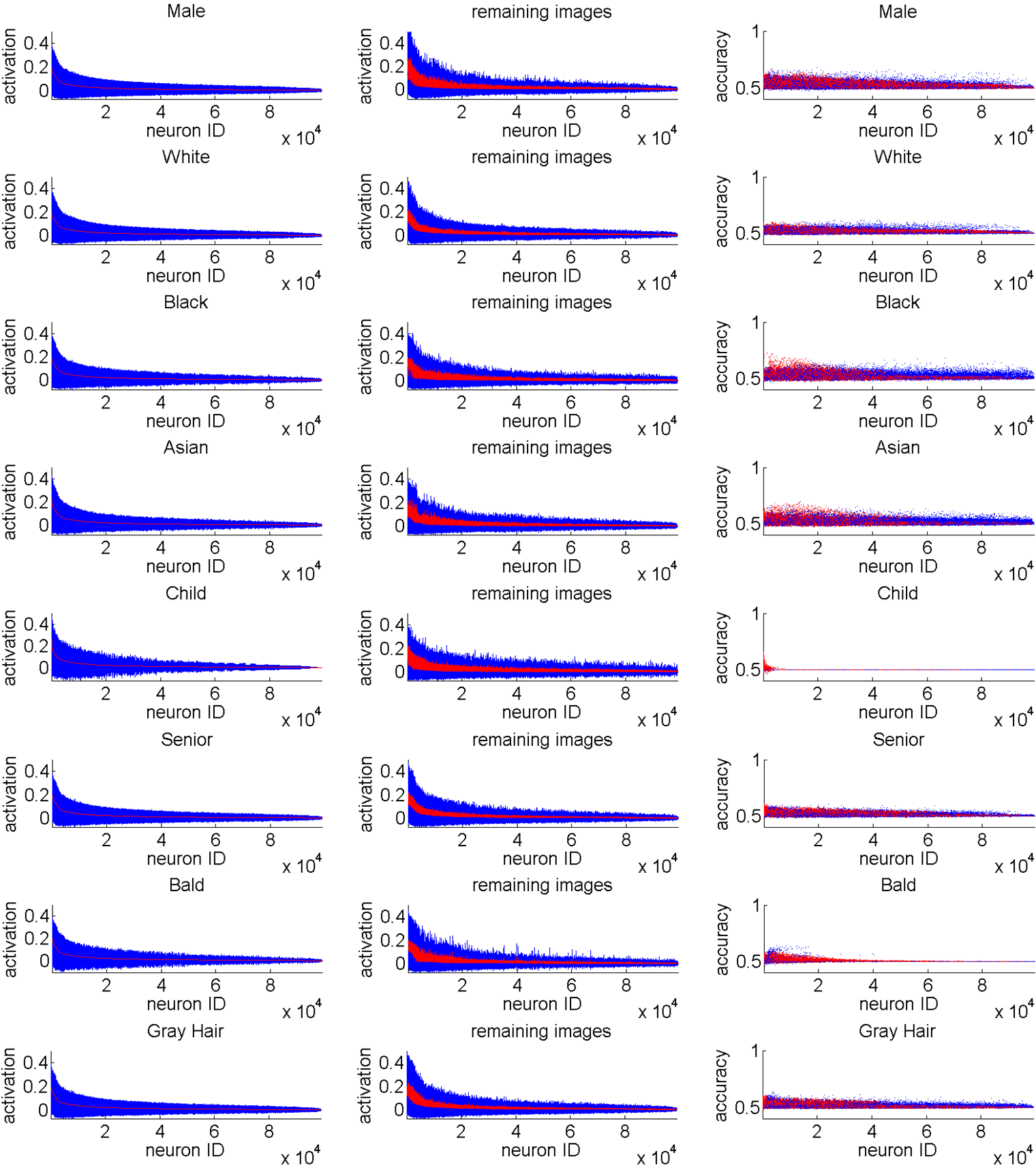}
\vspace{-0.02in}
\caption{LBP feature activation distributions and per-feature classification accuracies.}
\label{fig:attlbp}
\vspace{0.0in}
\end{subfigure}

\vspace{-0.02in}
\caption{Comparison of distributions of DeepID2+ neural and LBP feature activations and per-neuron (feature) classification accuracies of face images of particular attributes in LFW. Figure description is the same as Fig. \ref{fig:act}.}
\vspace{-0.05in}
\label{fig:att}
\end{figure}

Fig. \ref{fig:attcnn} compares the range of neural activations on faces containing a particular attribute (left column) and the remaining images (middle column), as well as showing the per-neuron classification accuracies of distinguishing each attribute from the remaining images (right column). Similar to identities, neurons of lower and higher ranks exhibit selectiveness to attributes as shown in this figure, including Male, White, Black, Asian, Child, Senior, Bald, and Gray Hair. These attributes are discriminative to identities. The selectiveness is relatively weak to other attributes such as Indian, Youth, Middle Aged, Black Hair, Blond Hair, and Brown Hair (not shown). These attributes are either visually ambiguous or less discriminative to identities. For example, Indian people sometimes look like Asian, and often times we see the same identity photographed at both youth and middle aged, or photographed in different hair colors.

Fig. \ref{fig:attlbp} compares the range of LBP features and per-feature classification accuracies for the same set of attributes as in Fig. \ref{fig:attcnn}. The range of LBP features on given attributes and the remaining images are overlapped for all features, and the classification accuracies are accumulated on the $50\%$ random guess line.

\subsection{Neural activation distribution}
\begin{figure}[t]
\begin{center}
\includegraphics[width=0.93\linewidth]{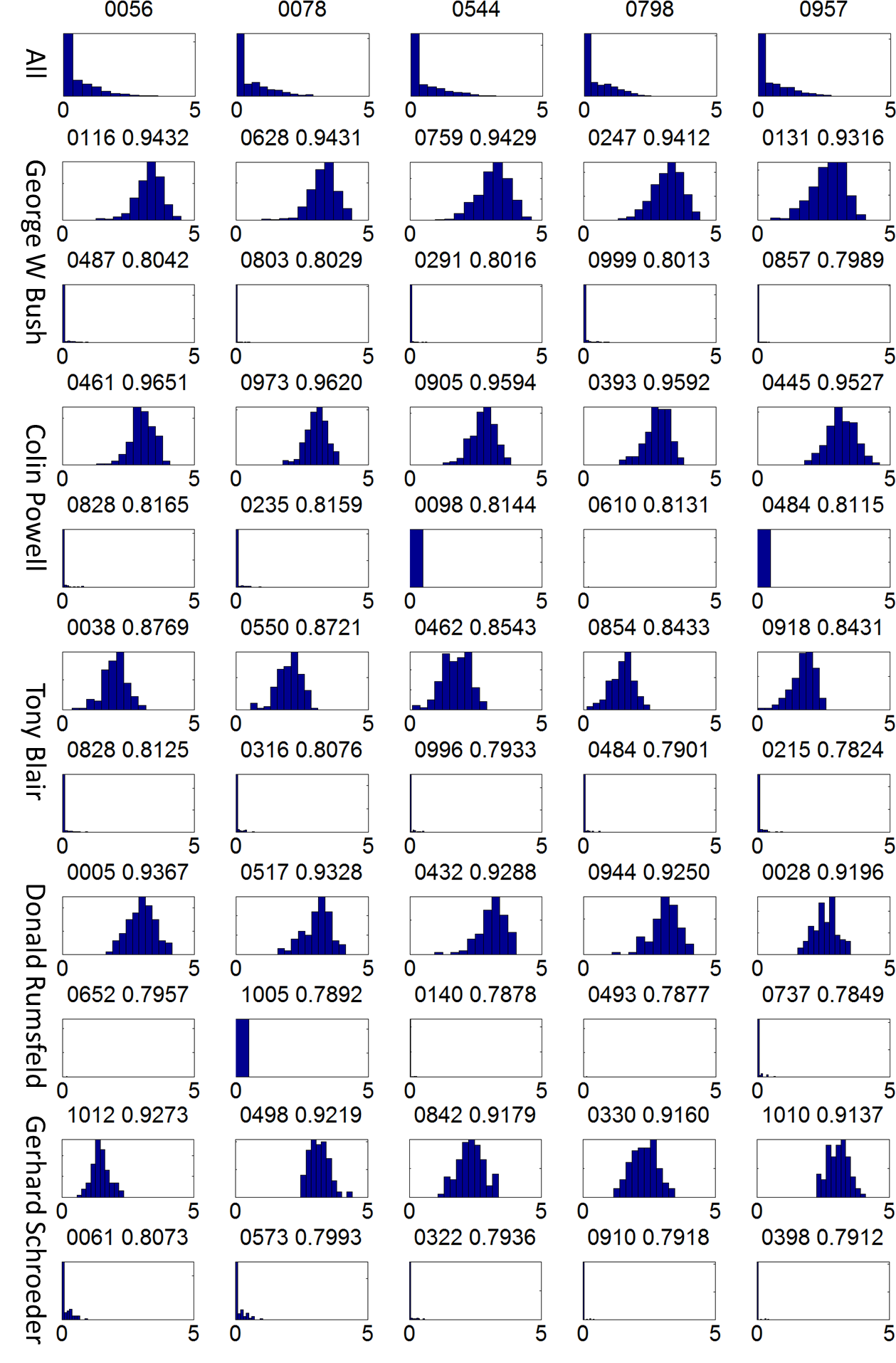}
\end{center}
\vspace{-0.1in}
\caption{Histogram of neural activations. First row: activation histograms over all face images of five randomly selected neurons, with neural ID labeled above each histogram. Second to the last row: activation histograms over the first five people with the most face images in LFW. For each person, histograms of five excitatory neurons (even rows) and five inhibitory neurons (odd rows except the first row) with the highest binary classification accuracies of distinguishing the given identity and the remaining images are shown. People names are given in the left of every two rows. Neural ID and classification accuracies are shown above each histogram.}
\label{fig:acthista}
\vspace{-0.05in}
\end{figure}

\begin{figure*}[tb!]
\centering

\begin{subfigure}{0.98\textwidth}
\includegraphics[width = 0.98\textwidth]{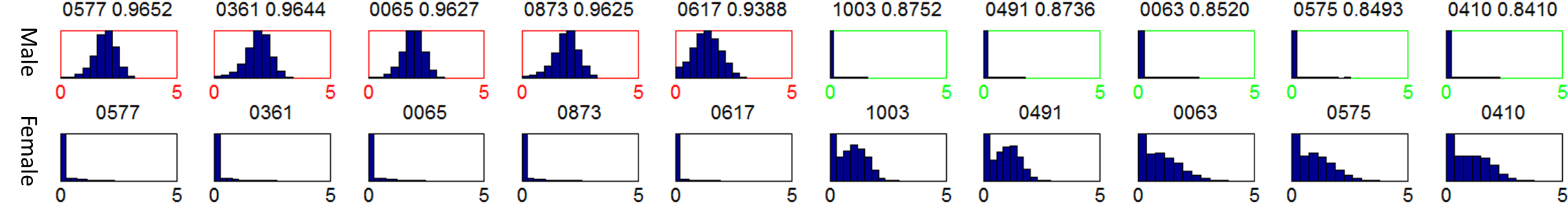}
\vspace{-0.06in}
\caption{Histogram of neural activations over sex-related attributes (Male and Female).}
\label{fig:sex}
\vspace{0.05in}
\end{subfigure}

\begin{subfigure}{0.88\textwidth}
\includegraphics[width = 0.88\textwidth]{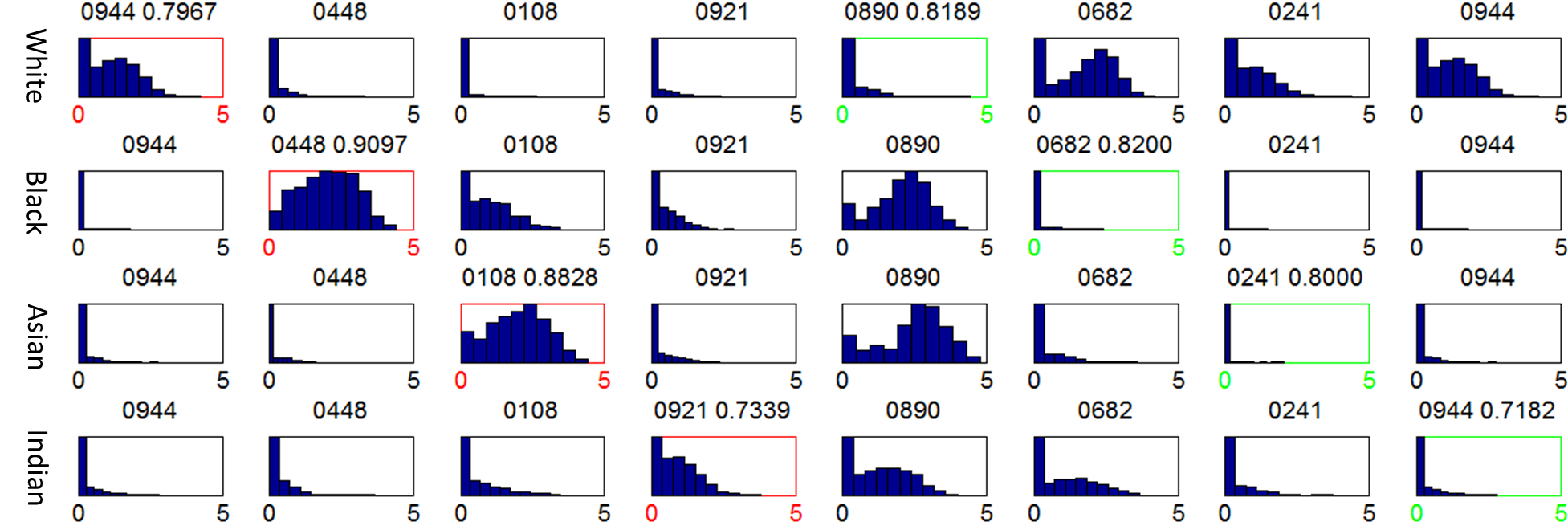}
\vspace{-0.06in}
\caption{Histogram of neural activations over race-related attributes, i.e., White, Black, Asian, and Indian.}
\label{fig:race}
\vspace{0.05in}
\end{subfigure}

\begin{subfigure}{0.98\textwidth}
\includegraphics[width = 0.98\textwidth]{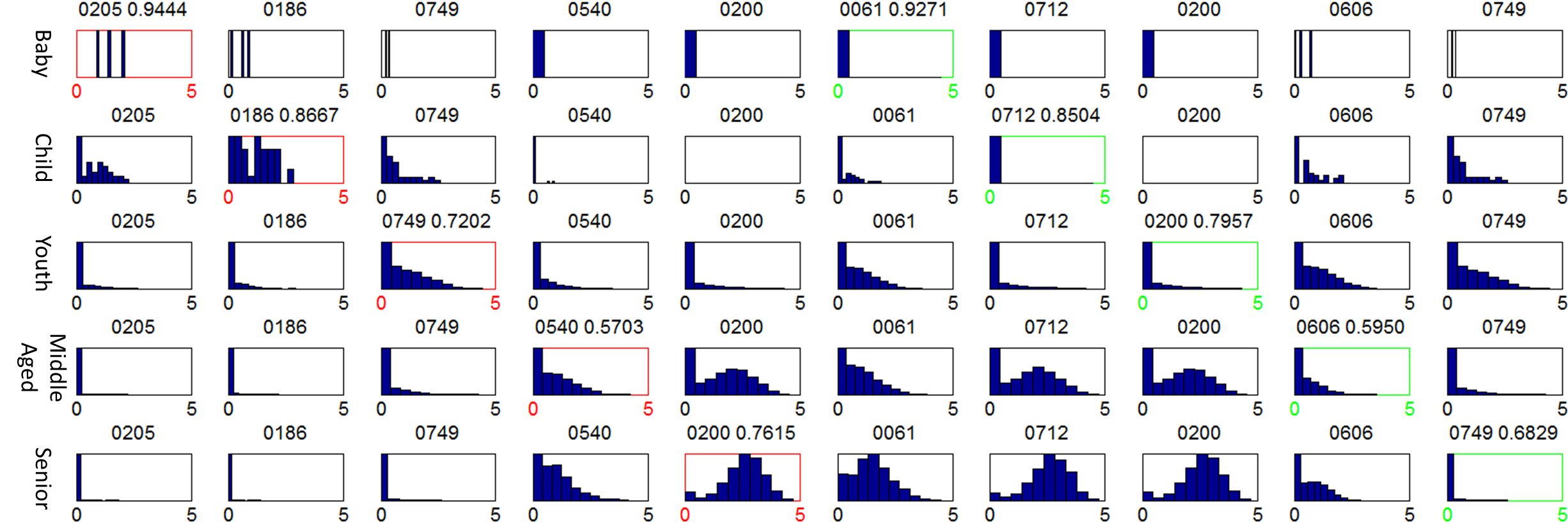}
\vspace{-0.06in}
\caption{Histogram of neural activations over age-related attributes, i.e., Baby, Child, Youth, Middle Aged, and Senior.}
\label{fig:age}
\vspace{0.05in}
\end{subfigure}

\begin{subfigure}{0.98\textwidth}
\includegraphics[width = 0.98\textwidth]{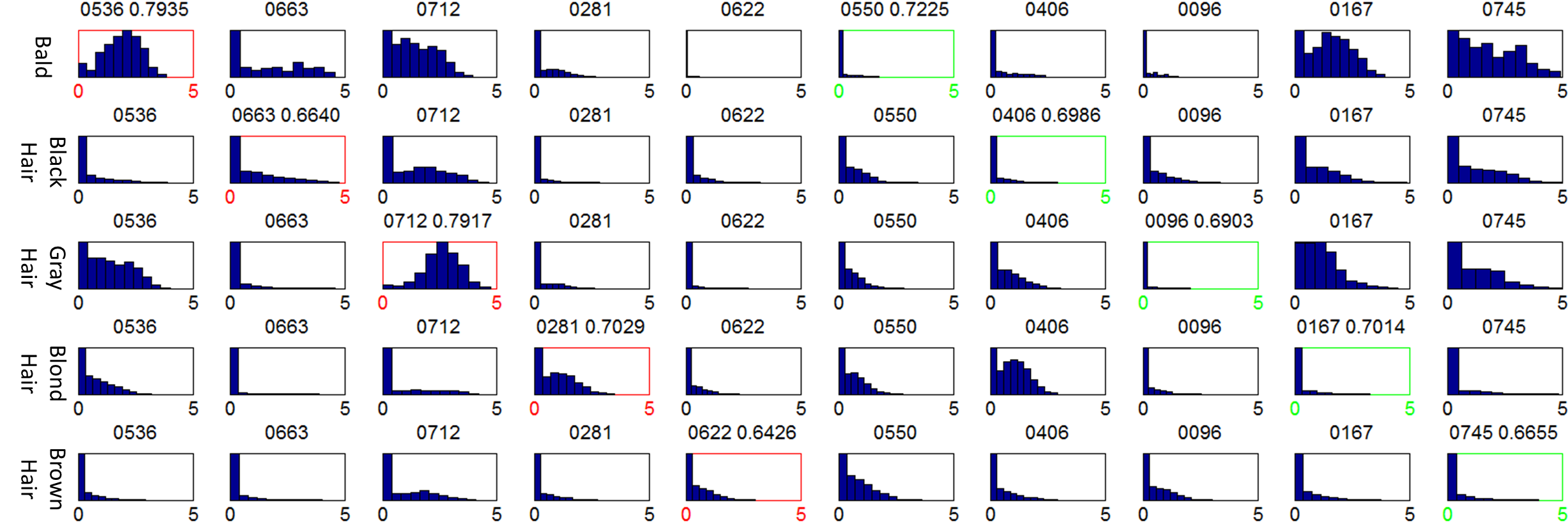}
\vspace{-0.06in}
\caption{Histogram of neural activations over hair-related attributes, i.e., Bald, Black Hair, Gray Hair, Blond Hair, and Brown Hair.}
\label{fig:hair}
\vspace{0.0in}
\end{subfigure}

\vspace{-0.05in}
\caption{Histogram of neural activations over attributes. Each column of each subfigure shows histograms of a single neuron over each of the attributes given in the left, respectively. Histograms of excitatory and inhibitory neurons which best distinguish each attribute from the remaining images are shown, and are framed in red and green, respectively, with neural ID and classification accuracies shown above each histogram. The other histograms are framed in black with only neural ID above.}
\vspace{-0.05in}
\label{fig:attr}
\end{figure*}

Fig. \ref{fig:acthista} and Fig. \ref{fig:attr} show examples of the histograms of neural activations over given identities or attributes. Fig. \ref{fig:acthista} first row also shows the histograms over all images of five randomly selected neurons. For each neuron, approximately half of its activations are zero (or close to zero) and another half have larger values. In contrast, the histograms over given identities exhibit strong selectiveness. Some neurons are constantly activated for a given identity, with activation histograms distributed in positive values, as shown in the first row of histograms of each identity in Fig. \ref{fig:acthista}, while some others are constantly inhibited, with activation histograms accumulated at zero or small values, as shown in the second row of histograms of each identity.

For attributes, in each column of Fig. \ref{fig:sex}, \ref{fig:race}, \ref{fig:age}, and \ref{fig:hair}, we show histograms of a single neuron over a few related attributes, \ie, those related to sex, race, age, and hair, respectively. The neurons are selected to be excitatory (in red frames) or inhibitory (in green frames) and can best classify the attribute shown in the left of each row. As shown in these figures, neurons exhibit strong selectiveness (either activated or inhibited) to certain attributes, in which the neurons are activated (inhibited) for the given attribute while inhibited (activated) for the other attributes in the same category.

\section{Robustness of DeepID2+ features}
\label{sec:robustness}

\begin{figure}[t]
\begin{center}
\includegraphics[width=0.9\linewidth]{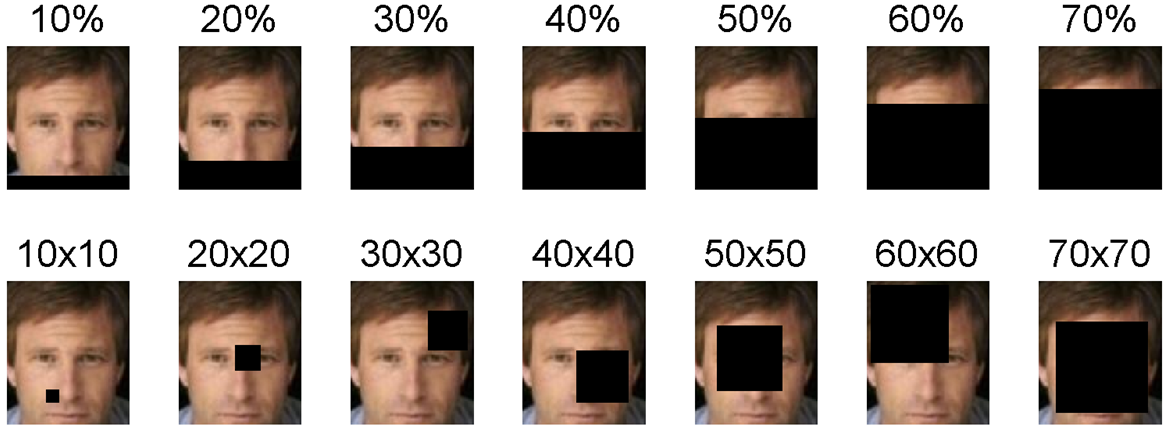}
\end{center}
\vspace{-0.15in}
\caption{The occluded images tested in our experiments. First row: faces with $10\%$ to $70\%$ areas occluded, respectively. Second row: faces with $10\times10$ to $70\times70$ random block occlusions, respectively.}
\label{fig:img_occlu}
\vspace{-0.1in}
\end{figure}

\begin{figure}[t]
\begin{center}
\includegraphics[width=0.9\linewidth]{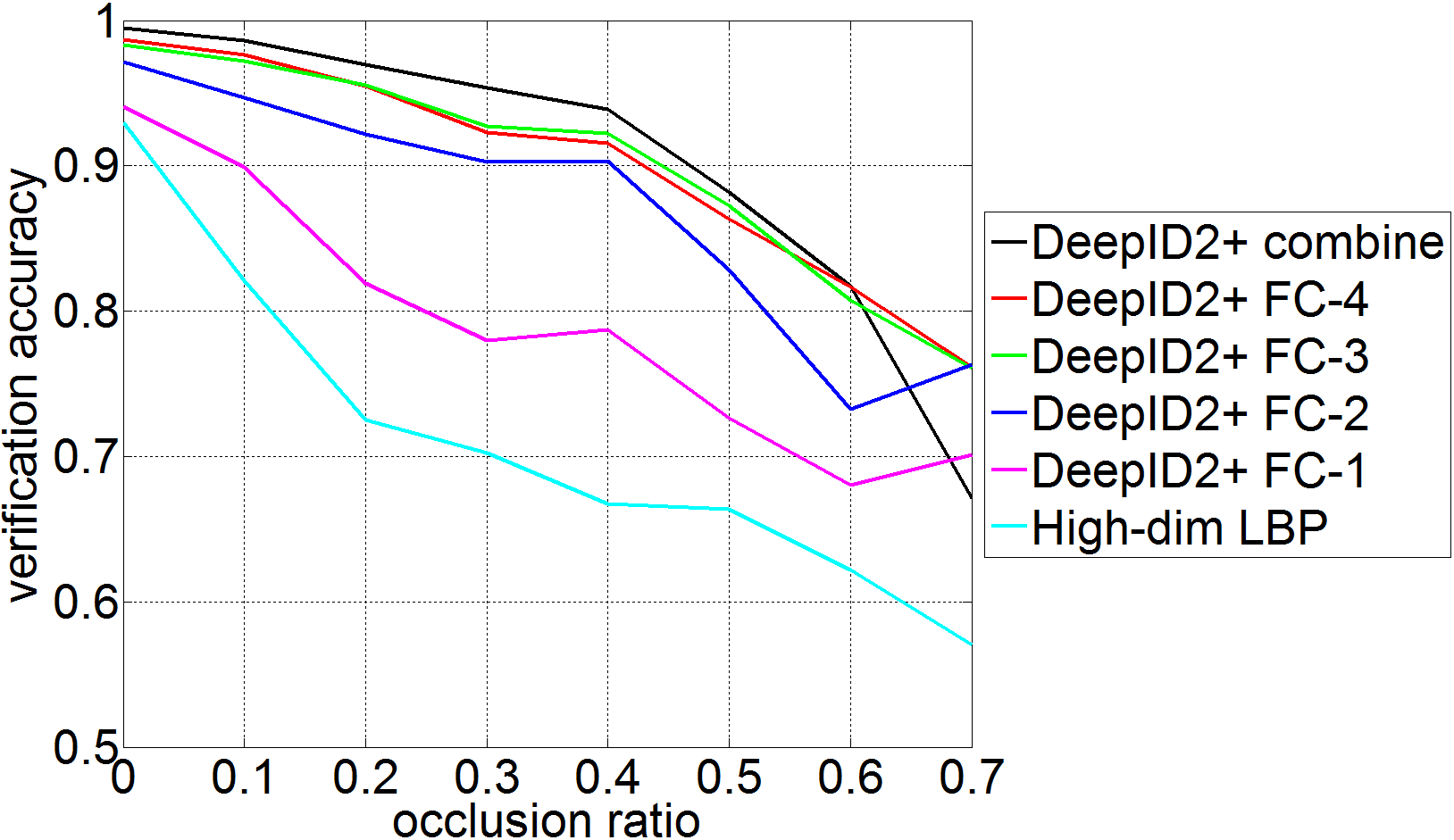}
\end{center}
\vspace{-0.15in}
\caption{Face verification accuracies of DeepID2+ and high-dimensional LBP on LFW with partial occlusions. The red, green, blue, and magenta curves evaluate the features of a single DeepID2+ net, extracted from various network depth (from FC-$4$ to FC-$1$ layer). We also evaluate the combination of $25$ DeepID2+ net FC-4 layer features, shown by the black curve.}
\label{fig:accu_occlu}
\vspace{-0.15in}
\end{figure}

We test the robustness of DeepID2+ features on face images with occlusions. In the first setting, faces are partially occluded by $10\%$ to $70\%$ areas, as shown in Fig. \ref{fig:img_occlu} first row. In the second setting, faces are occluded by random blocks of $10\times10$ to $70\times70$ pixels in size, as shown in Fig. \ref{fig:img_occlu} second row. In the occlusion experiments, DeepID2+ nets and Joint Bayesian models are learned on the original face images in our training set without any artificially added occlusions, while the occluded faces are only used for test. We also test the high-dimensional LBP features plus Joint Bayesian models \cite{chen2013} for comparison. Fig. \ref{fig:accu_occlu} compares the face verification accuracies of DeepID2+ and LBP features on LFW test set with varying degrees of partial occlusion. The DeepID2+ features are taken from the FC-$1$ to FC-$4$ layers with increasing depth in a single DeepID2+ net taking the entire face region as input. We also evaluate our entire face recognition system with $25$ DeepID2+ nets. The high-dimensional LBP features compared are $99,120$ dimensions extracted from $21$ facial landmarks. As shown in Fig. \ref{fig:accu_occlu}, the performance of LBP  drops dramatically, even with slight $10\%$ - $20\%$ occlusions. In contrast, for the DeepID2+ features with two convolutions and above (FC-2, FC-3, and FC-4), the performance degrades slowly in a large range. Face verification accuracies of DeepID2+ are still above $90\%$ when $40\%$ of the faces are occluded (except FC-$1$ layer), while the performance of LBP features has dropped below $70\%$. The performance of DeepID2+ only degrades quickly with over $50\%$ occlusions, when the critical eye regions are occluded. It also shows that features in higher layers (which are supposed to be more globally distributed) are more robust to occlusions, while both LBP and FC-$1$ are local features, sensitive to occlusions. Combining DeepID2+ nets extracted from $25$ face regions achieves the most robustness with $93.9\%$ face verification accuracy with $40\%$ occlusions and $88.2\%$ accuracy even only showing the forehead and hairs.

\begin{figure}[t]
\begin{center}
\includegraphics[width=0.9\linewidth]{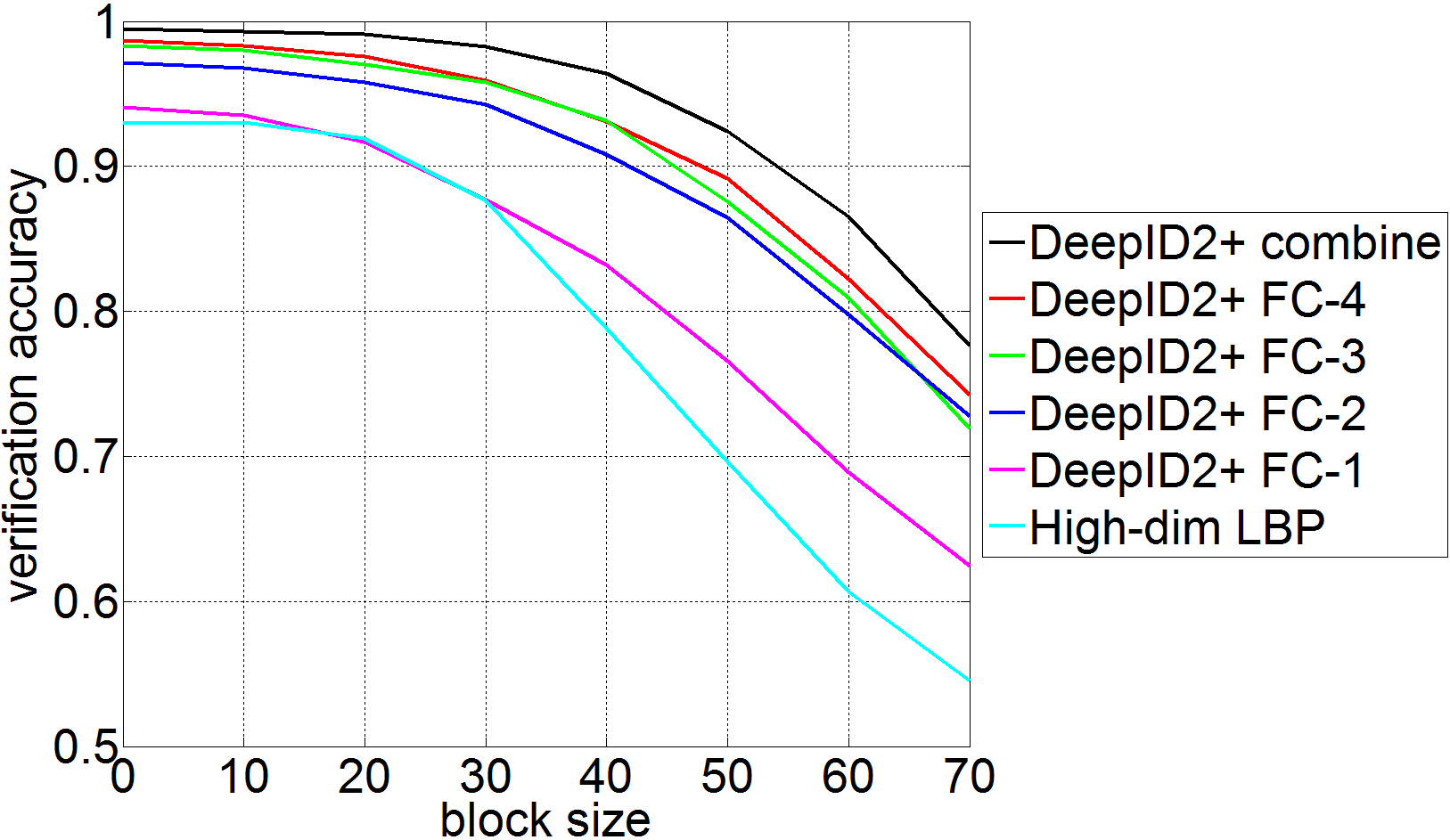}
\end{center}
\vspace{-0.15in}
\caption{Face verification accuracies of DeepID2+ and high-dimensional LBP on LFW with random block occlusions. Curve description is the same as Fig. \ref{fig:accu_occlu}.}
\label{fig:accu_rb}
\vspace{-0.15in}
\end{figure}

We also evaluate face verification of DeepID2+ and LBP features over face images with random block occlusions, with $n\times n$ block size for $n=10$ to $70$, respectively. This setting is challenging since the positions of the occluded regions in two faces to be verified are generally different. Therefore images of the same person would look much different in the sense of pixel distances. Fig. \ref{fig:accu_rb} shows the comparison results, the accuracies of LBP features begin to drop quickly when block sizes are greater than $20\times20$, while DeepID2+ features (except FC-$1$) maintain the performance in a large range. With $50\times50$ block occlusions, the performance of LBP features has dropped to approximately $70\%$, while the FC-$4$ layer of a single DeepID2+ net still has $89.2\%$ accuracy, and the combination of $25$ DeepID2+ nets has an even higher $92.4\%$ accuracy. Again, the behavior of features in the shallow FC-$1$ layer are closer to LBP features. The above experiments show that it is the deep structure that makes the neurons more robust to image corruptions. Such robustness is inherent in deep ConvNets without explicit modelings.

\begin{figure}[t]
\begin{center}
\includegraphics[width=0.95\linewidth]{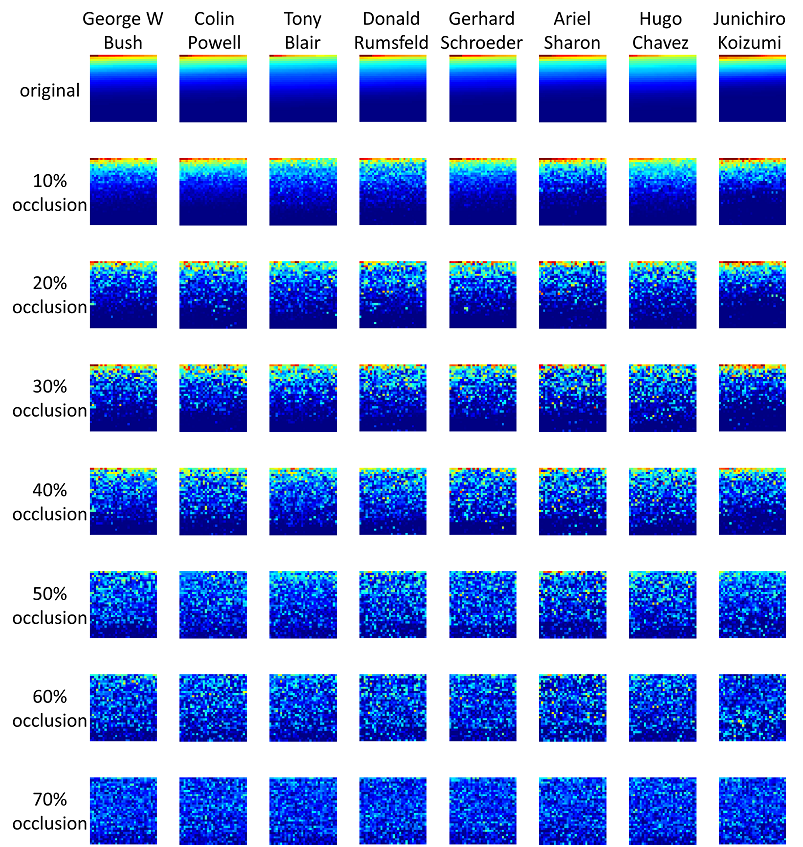}
\end{center}
\vspace{-0.1in}
\caption{Mean neural activations over partially occluded face images (shown in Fig. \ref{fig:img_occlu} first row). Each column shows the mean activations over face images of a single identity given in the top of each column, with various degrees of occlusions given in the left of each row. Neurons in figures in each column are sorted by their mean activations on the original images of each identity. Activation values are mapped to a color map with warm colors indicating positive values and cool colors indicating zero or small values.}
\label{fig:act_occlu_part}
\vspace{-0.05in}
\end{figure}

\begin{figure}[t]
\begin{center}
\includegraphics[width=0.95\linewidth]{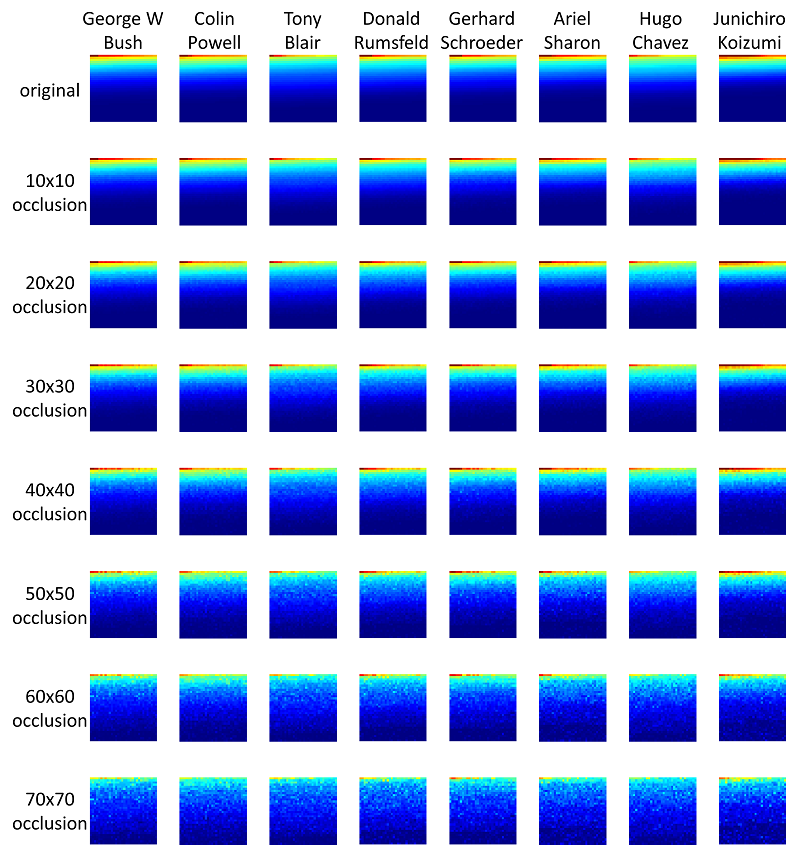}
\end{center}
\vspace{-0.1in}
\caption{Mean neural activations over images with random block occlusions (shown in Fig. \ref{fig:img_occlu} second row). Figure description is the same as Fig. \ref{fig:act_occlu_part}.}
\label{fig:act_occlu_rb}
\vspace{-0.05in}
\end{figure}

Fig. \ref{fig:act_occlu_part} and \ref{fig:act_occlu_rb} show the mean activations of FC-$4$ layer neurons over images of a single identity with various degrees of partial and random block occlusions, respectively. The neurons are ordered according to their mean activations on the original images of each identity. For both types of occlusions, activation patterns keep largely unchanged until a large degree of occlusions.

\section{Conclusion}

This paper designs a high-performance DeepID2+ net which sets new sate-of-the-art on LFW and YouTube Faces for both face identification and verification. Through  empirical studies, it is found that the face representations learned by DeepID2+ are moderately sparse, highly selective to identities and attributes, and robust to image corruption. In the past, many research works have been done aiming to achieve such attractive properties by explicitly adding components or regularizations to their models or systems. However, they can be naturally achieved by the deep model through large scale training. This work not only significantly advances the face recognition performance, but also provides valuable insight to help people to understand deep learning and its connection with many existing computer vision researches such as sparse representation, attribute learning and occlusion handling. Such insights may inspire  more exciting research in the future. As an example, in this work, we have shown that binary neural activation patterns are highly efficient and effective for face recognition.

{\small
\bibliographystyle{ieee}
\bibliography{deepid2plus}
}

\end{document}